\documentclass[preprint]{elsarticle}
\usepackage{comment,lineno,hyperref,natbib,amssymb,multirow,booktabs,subfig,color,tabu,arydshln}
\usepackage[bottom]{footmisc}

\journal{Medical Image Analysis}

\newcommand{\tabincell}[2]{\begin{tabular}{@{}#1@{}}#2\end{tabular}}

\biboptions{authoryear}

\begin{document}

\begin{frontmatter}

\title{Neural Multi-Atlas Label Fusion: Application to Cardiac MR Images}

\author[address1]{Heran Yang}

\author[address1,address2]{Jian Sun\corref{corrauthor}}
\cortext[corrauthor]{Corresponding author}
\ead{jiansun@mail.xjtu.edu.cn}

\author[address1,address2]{Huibin Li}

\author[address3]{Lisheng Wang}

\author[address1,address2]{Zongben Xu}

\address[address1]{School of Mathematics and Statistics, Xi'an Jiaotong University, China}
\address[address2]{National Engineering Laboratory for Big Data Algorithm and Analysis Technology, China}
\address[address3]{Department of Automation, Shanghai Jiaotong University, China}

\begin{abstract}
Multi-atlas segmentation approach is one of the most widely-used image segmentation techniques in biomedical applications. There are two major challenges in this category of methods, \emph{i.e.}, atlas selection and label fusion. In this paper, we propose a novel multi-atlas segmentation method that formulates multi-atlas segmentation in a deep learning framework for better solving these challenges. The proposed method, dubbed deep fusion net (DFN), is a deep architecture that integrates a feature extraction subnet and a non-local patch-based label fusion (NL-PLF) subnet in a single network. The network parameters are learned by end-to-end training for automatically learning deep features that enable optimal performance in a NL-PLF framework. The learned deep features are further utilized in defining a similarity measure for atlas selection.
By evaluating on two public cardiac MR datasets of SATA-13 and LV-09 for left ventricle segmentation, our approach achieved 0.833 in averaged Dice metric (ADM) on SATA-13 dataset and 0.95 in ADM for epicardium segmentation on LV-09 dataset, {comparing} favorably with the other automatic left ventricle segmentation methods. We also tested our  approach on Cardiac Atlas Project (CAP) testing set of MICCAI 2013 SATA Segmentation Challenge, and our method achieved 0.815 in ADM, ranking highest at the time of writing.
\end{abstract}


\end{frontmatter}


\section{Introduction}
As one of the most successful medical image segmentation techniques, multi-atlas segmentation (MAS) approach has been applied to various medical image segmentation tasks, including segmentation of 
abdominal anatomy~\citep{Tong2015,Wang2014,Wolz2013,Xu2015}, 
cardiac ventricle~\citep{Xie2015,Bai2015,Bai2013,Zhuang2016}, brain~\citep{Wu2014,Asman2013,Coupe2011,Albert2013,Artaechevarria2009,Sanroma2014,Wang2013,Cardoso2013,Asman2015}, etc.

Given an image to be segmented, \emph{i.e.}, a \emph{target image}, multi-atlas segmentation methods utilize multiple \emph{atlases}, \emph{i.e.}, a number of images from multiple subjects with segmentation labels delineated by experts, to estimate the segmentation label of target image, \emph{i.e.}, \emph{target label}. Typically,  multi-atlas segmentation methods first register atlas images to the target image, and then the corresponding warped atlas labels are combined to estimate the target label by a label fusion procedure~\citep{Iglesias2015}. To raise computational efficiency or improve final segmentation accuracy, multi-atlas segmentation methods employ an atlas selection procedure to select a few warped atlas images most similar to the target image, and only the labels of these selected images are utilized in the label fusion procedure~\citep{Aljabar2009}. For a comprehensive review on multi-atlas segmentation methods, please refer to \cite{Iglesias2015}.

A large body of literatures on multi-atlas segmentation {focuses} on the label fusion procedure. 
One typical label fusion strategy is weighted voting, 
where the label of each target voxel is determined by weighted average of the labels of corresponding voxels in warped atlas images.
Local label fusion methods determine the voxel-wise fusion weights by local intensity-based similarities between the target and atlas voxels~\citep{Artaechevarria2009,Makropoulos2014}.
To account for possible registration errors, \cite{Coupe2011} proposed to fuse labels of all voxels in a non-local search window or volume around the registered atlas voxel for predicting the target label.  {This category of methods~\citep{Bai2015,Iglesias2015,Wang2014} is commonly named as \textit{non-local patch-based label fusion}}. { Moreover, statistical fusion methods were proposed to estimate the fusion weights by integrating models of rater performance~\citep{Cardoso2013,Asman2013,Warfield2004}.
Instead of using weighted voting strategy, sparsity-based dictionary learning~\citep{Tong2015} and matrix completion~\citep{Sanroma2015} methods predict target labels by representing image and label patch pairs using sparse regularization or low rank constraint. These related methods commonly use intensities or hand-crafted features for representing atlas and target voxels / patches to measure atlas-to-target similarity.  A fundamental question is that whether we can automatically learn the features of atlas and target images directly aiming at achieving optimal label fusion performance.}

Atlas selection, another important issue in multi-atlas segmentation, aims at selecting the most relevant atlases, which is generally achieved by ranking the atlases according to their similarities to the target image. 
The traditional methods rank atlases using intensity-based similarity measures, \emph{e.g.}, normalized mutual information~\citep{Studholme1999,Collignon1995}. Other well designed measures for atlas selection include the distance between transformations~\citep{Commowick2009}, registration consistency~\citep{Heckemann2009}, \emph{etc}.
Machine learning methods were also introduced to learn similarity measures for atlas selection, \emph{e.g.}, manifold learning~\citep{Albert2013}, ranking SVM~\citep{Sanroma2014}, \emph{etc}.  For atlas selection, how to design image features to measure the atlas-to-target image similarity for relevant  atlas selection is also a fundamental task.

This work aims at designing a  feature learning-based approach for better achieving label fusion and atlas selection in multi-atlas segmentation. We propose a novel multi-atlas segmentation method by reformulating non-local patch-based label fusion (NL-PLF) method~\citep{Coupe2011,Bai2015}  to be a deep neural network~\citep{Krizhevsky2012}.   As shown in Fig.~\ref{fig:DFN_sturcture}, the network is comprised of a \textit{feature extraction subnet} for extracting deep features of each voxel in atlas and target images, and a \textit{non-local patch-based label fusion subnet} (NL-PLF subnet) for fusing the warped atlases based on the extracted deep features.  Our proposed deep fusion net relies on atlas-to-target image registration, and the net concentrates on learning deep features for fusing these warped labels using NL-PLF by an end-to-end training strategy. More specifically, the feature extraction subnet is learned to embed image into a deep feature space in which the feature similarity well reflects the label similarity,  and then these deep features are utilized in NL-PLF subnet to compute the label fusion weights for achieving optimal segmentation performance in a NL-PLF framework. The learned deep features for label fusion can also be taken as the features for measuring atlas-to-target image similarity in atlas selection.

To the best of our knowledge, this is the first work accomplishing registration-based multi-atlas segmentation in a deep learning framework.  
The traditional non-local patch-based label fusion method relies on hand-crafted features, \emph{e.g.}, intensities~\citep{Coupe2011}, contextual features \citep{Bai2015}, for computing label fusion weights.  Our approach
takes advantage of the strong feature learning ability of deep neural network to learn optimal image features for label fusion in multi-atlas segmentation. This is achieved by our specially designed network architecture integrating the modules of feature learning and label fusion in an end-to-end learning framework.

We apply the proposed deep fusion net to left ventricle (LV) segmentation from {short-axis} cardiac MR images~\citep{Petitjean2011}, and achieved competitive  results  on two publicly available cardiac MR datasets, \emph{i.e.}, MICCAI 2013 SATA Segmentation Challenge (SATA-13) dataset~\footnote{{https://www.synapse.org/\#!Synapse:syn3193805/wiki/217780}}~\citep{Bennett2013} and MICCAI 2009 LV Segmentation Challenge (LV-09) dataset~\footnote{https://smial.sri.utoronto.ca/LV\_Challenge/Home.html}~\citep{Radau2009}. Our approach achieved 0.833 in averaged Dice metric (ADM) on SATA-13 dataset and 0.95 in ADM for epicardium segmentation on LV-09 dataset. 
On the Cardiac Atlas Project (CAP) testing set of MICCAI 2013 SATA Segmentation Challenge, our method achieved 0.815 in Dice metric, ranking highest on this dataset at the time of writing.

Note that a preliminary version of this work was published in \cite{Yang2016}.
Compared with the conference paper, this journal paper presents the following extensions. 
(1) We more comprehensively discuss on the motivations and details of the proposed method for better readability.
(2) Our method is evaluated on additional dataset (LV-09 dataset) and compared with more segmentation methods, including the state-of-the-art deep learning methods.
(3) More experiments are conducted to further explore the behaviors of the proposed method, \emph{e.g.}, impact of atlas selection strategy, cross-dataset evaluation, \emph{etc}.

\subsection{Related work}

\subsubsection{Deep learning approach in medical image segmentation}
In recent years, deep learning approach was widely applied in medical image segmentation, \emph{e.g.}, vessel segmentation~\citep{Fu2016,Maninis2016,Khalaf2016}, brain segmentation~\citep{Wachinger2017,Moeskops2016,Rajchl2017}, \emph{etc}. 
These methods commonly design different deep network structures considering backgrounds of specific problems, and directly learn the optimal network parameters for voxel-wise label prediction. 
For example, \cite{Havaei2017} proposed two-pathway cascaded deep networks with a two-phase training procedure for  brain tumor segmentation. 
\cite{Chen2017} designed a multi-task fully convolutional neural network with multi-level contextual information for object instance segmentation from histology images.
{\cite{Ronneberger2015} proposed a data augmentation strategy by applying elastic deformations to the available training images, and designed a U-net architecture that consists of a contracting path to capture context and a symmetric expanding path enabling precise localization for cell segmentation in microscopic images.}
Moreover, \cite{Kamnitsas2017} proposed a multi-scale 3D convolutional neural network with two convolutional pathways for lesion segmentation in multi-modal brain MRI.
In addition, some methods concentrate on defining proper loss functions for training deep segmentation networks on medical images.
For example, \cite{Milletari2016} proposed a novel loss function based on Dice coefficient for prostate segmentation to deal with the strong imbalance between the number of foreground and background voxels.
\cite{BenTaieb2016} designed a new loss for segmentations of histology glands to encode geometric and topological priors of containment and detachment.
For a comprehensive review on deep learning in medical image analysis, please refer to \cite{litjens2017}.

\begin{figure*}
	\centering
	\includegraphics[width=12.2cm]{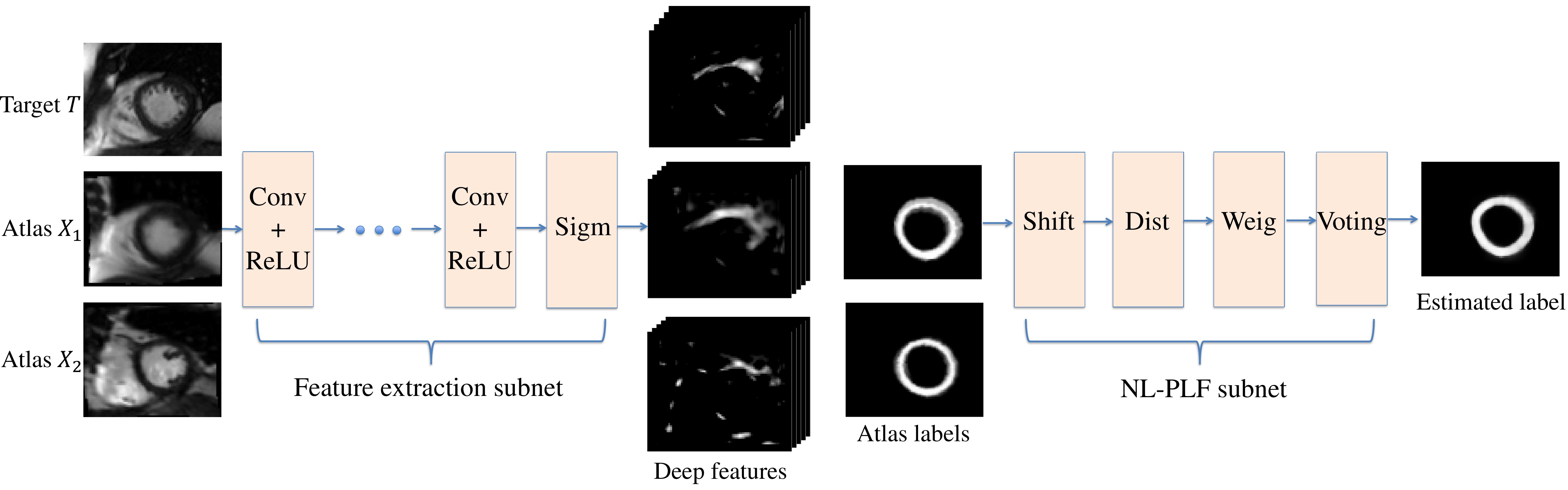}
	\caption{The architecture of deep fusion net. A target image $T$ and its warped atlas images $X_i$ are first fed into a {\em feature extraction subnet} to extract deep features. These features and atlas labels are then sent to a {\em non-local patch-based label fusion (NL-PLF) subnet} for generating the estimated label of target image,  {\em i.e.}, target label.}
	\label{fig:DFN_sturcture}
\end{figure*}

\subsubsection{Cardiac MR left ventricle segmentation}
Accurately segmenting the LV from cardiac MR images is essentially important for quantitatively assessing the cardiac function in clinical diagnosis, which is still a challenging task due to large variation of LV in intensity levels, structural shapes, respiratory motion artifacts, partial volume effects, \emph{etc}.

Multi-atlas segmentation approach is widely applied in LV segmentation. \cite{Bai2015} incorporated intensity, gradient and contextual information into an augmented feature vector for similarity measure in the non-local patch-based label fusion framework.
Instead of only considering the corresponding atlas slice to the target image, \cite{Xie2015} fused the labels of all atlas slices in a neighborhood to produce a more accurate estimated target label. 
\cite{Wang2013} formulated the weighted voting as a problem of minimizing total expectation of labeling error, and pairwise dependency between atlases was modeled as a joint probability of two atlases making a segmentation error at a voxel.
Due to the close connection between label fusion and registration, \cite{Bai2013} attempted to alternately perform patch-based label fusion and registration refinement.

In recent years, deep learning approach was also applied to LV segmentation. For example, \cite{Avendi2016}  utilized convolutional neural networks to determine the region of interest (ROI) containing LV and stacked autoencoders to infer the LV segmentation mask. Then the segmentation mask was  incorporated into level set model to produce the final segmentation result. 
\cite{Ngo2017} first estimated the ROI and an initial segmentation by deep belief network and Otsu's thresholding, and then refined the initial segmentation by level set model.
These two methods achieved high accuracies on MICCAI 2009 LV Segmentation Challenge dataset, but their performance largely relies on the post-processing by level set method. In these work, the segmentation results purely using deep networks are unsatisfactory~\citep{Avendi2016,Ngo2017,Guo2016} possibly due to  insufficient labeled training data or limitation of designed network architecture, and the post-processing by level set method helps to refine the segmentation results to be continuous and connected.

Our proposed deep fusion net bridges the multi-atlas segmentation approach and deep learning approach. Compared with the traditional multi-atlas segmentation methods, our net also relies on atlas-to-target image registration. But instead of using hand-crafted features for label fusion, our net learns to extract deep features for computing optimal fusion weights in a NL-PLF framework. Compared with the common deep learning methods directly learning mapping from image to semantic labels, our network architecture is non-conventional and inspired by a registration-based multi-atlas label fusion strategy. With this novel deep architecture, our approach achieved improved results compared with the traditional  multi-atlas segmentation methods and automatic deep learning methods on two cardiac MR image datasets, and the only deep learning method surpassing our results depends on a strong manual prior~\citep{Ngo2017}.

\section{Deep Fusion Net for Multi-Atlas Segmentation}
This section presents the proposed deep fusion net for multi-atlas segmentation, including the general framework, network architecture,  training and testing procedures.

\subsection{General framework}
\label{sec:framework}
As a registration-based multi-atlas segmentation method, multiple {atlas images} are first registered to the target image by a non-rigid registration method, and then the corresponding atlas labels are warped to the target image using these transformations. Our proposed deep fusion net  (DFN) is designed to fuse the warped atlas labels using discriminatively learned deep features. As shown in Fig.~\ref{fig:DFN_sturcture}, deep fusion net consists of a feature extraction subnet, followed by a non-local patch-based label fusion (NL-PLF) subnet.

\textbf{\textit{Feature extraction subnet:}} The feature extraction subnet is responsible for extracting dense deep features from target and atlas images, and all these images share the same feature extraction subnet. {This} subnet is learned to embed the input images into a deep feature space, in which the feature similarity between paired voxels are expected to better reflect the similarity of their  labels. The extracted dense deep features are utilized by the following NL-PLF subnet for computing the label fusion weights using the deep feature similarity.

\textbf{\textit{NL-PLF subnet:}} The NL-PLF subnet aims to fuse the warped atlas labels {for predicting the target label}.  It implements a non-local patch-based label fusion strategy that predicts {the} target label by the weighted average of atlas labels within a search volume around the registered voxels.  The effectiveness of this label fusion strategy relies on an effective measure of feature similarity for the fusion weight computation. This subnet takes the extracted deep features {by} feature extraction subnet as input, and accomplishes the non-local patch-based label fusion using the weights computed based on these extracted features.

The deep fusion net integrates the modules of feature extraction for feature embedding and label fusion for target label prediction. The network parameters are learned by an end-to-end training procedure minimizing a loss defined between network output and ground-truth target label.  By enforcing that the estimated segmentation {label} should approximate the target label in training, the feature extraction subnet is discriminatively learned to extract deep features with feature similarity well representing the label similarity for computing the optimal label fusion weight.

In computer vision, siamese neural networks were proposed for learning deep feature embedding, and the learned feature similarity was utilized for patch matching~\citep{Zagoruyko2015}, face recognition~\citep{Chopra2005,Schroff2015} and re-identification~\citep{Zheng2016}, \emph{etc}.  Similarly, our feature extraction subnet aims to embed image patches~\footnote{The image patch size is determined by the receptive field of feature extraction subnet.} into a deep feature space, in which the feature similarity is a meaningful measure of semantic label similarity.
The deep feature similarity is further utilized to define label fusion weights for warped atlas label fusion or retrieve relevant atlases in atlas selection.

In the following sections 2.2 and 2.3, we will respectively introduce the architectures of feature extraction subnet and NL-PLF subnet.

\begin{figure*}[tp]
	\centering
	\includegraphics[width=12.2cm]{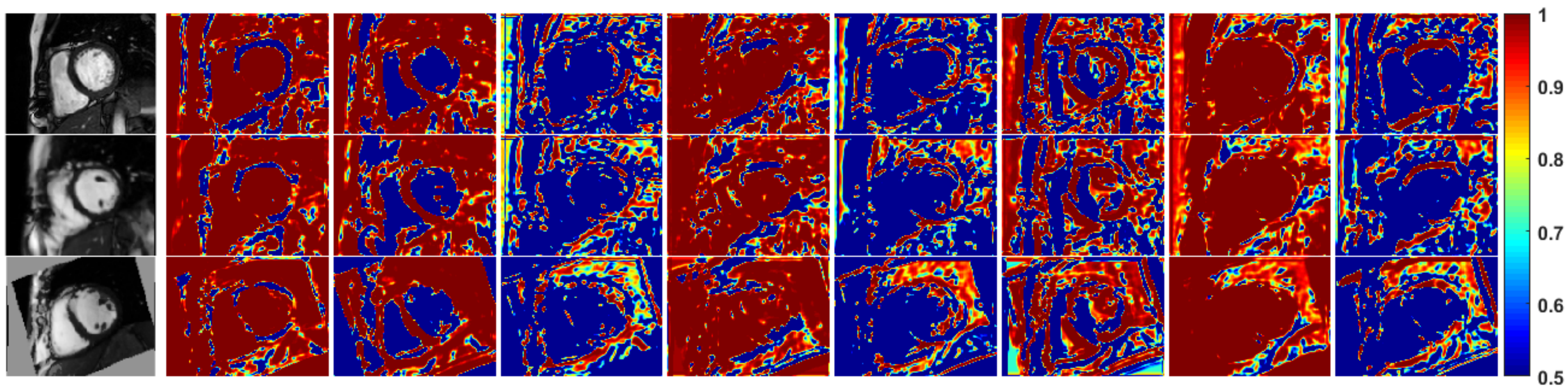}
	\caption{ Visualization of extracted features by learned feature extraction subnet. Top to bottom: a target image, two warped atlas images. Left to right: input image, first eight extracted feature maps.}
	\label{fig:DFN_extractedFeatures}
\end{figure*}

\subsection{Feature extraction subnet}
\label{feat_extract_subnet}
Assume that we are given a target image $T$ and multiple atlases. After registering atlas images to the target image $T$, the pairs of warped atlas image and label map are denoted as $\{ X_i,  L(X_i)\}_{i=1}^K$, where $X_i$ is the $i$-th warped atlas image, and $L(X_i)$ is the corresponding warped label map. {$K$ denotes the number of atlases.} The feature extraction subnet is responsible for extracting dense deep features from images, including the target image $T$ and warped atlas images $\{ X_i\}_{i=1}^K$, and the extracted dense features are respectively denoted as $F(T)$ and $\{F(X_i)\}_{i=1}^K$.

As shown in Fig.~\ref{fig:DFN_sturcture}, the feature extraction subnet consists of multiple repetitions of convolutional layer with ReLU activation function, and a final sigmoid layer for feature normalization.
All input images of this subnet, including the target image and warped atlas images, share the same subnet. {Figure~\ref{fig:DFN_extractedFeatures} shows examples of extracted feature maps of one target image and two warped atlas images by the feature extraction subnet after training. These extracted features well describe different structures of the input images.  }

\textit{\textbf{Convolutional block.}}
The convolutional block aims at learning discriminative local patterns represented by filters from its input features.
This block convolves the input features using a set of learnable filters $\{ \mathcal{W}_{d'} \}_{d'=1}^{D'}$, followed by non-linear activation function. Each filter $\mathcal{W}_{d'} \in \mathbb{R}^{w_f \times w_f \times D}$ is a third-order tensor, where $D'$ is the number of filters, $D$ denotes the number of \emph{feature maps} in input features, and $w_f \times w_f$ is the size of filters. 
Given the input features $G^{l-1}({X}) \in \mathbb{R}^{M \times N \times D}$ of image $X$, the $l-$th convolutional block outputs features $G^{l}({X}) \in \mathbb{R}^{(M - w_f + 1) \times (N - w_f + 1) \times D'}$, 
whose $d'$-th feature map, denoted as ${g}^l_{d'}$, can be written as 
\begin{equation}
	{g}^l_{d'} = \varphi \left(  {\mathcal{W}}_{d'} * G^{l-1}(X) + b_{d'}^l \right),
\end{equation}
where $*$ is the 3-D convolution operator, and $b_{d'}^l$ is the bias term. 
$\varphi$ is a rectified linear unit (ReLU) function~{\citep{Nair2010}}: $\varphi(x) = \max (0, x) $.

\textit{\textbf{Sigmoid layer.}}
The feature extraction subnet is {completed with a final} sigmoid layer. The sigmoid layer can suppress the large feature magnitude and non-linearly  map the extracted features to a limited range within $[0.5, 1)$, which is expected to enforce robustness of deep fusion net to different image contrasts. The experimental evaluation on the necessity of this sigmoid layer is presented in section~\ref{sec:nece_sigmoid_layer}.
Given input features $G({X})  \in \mathbb{R}^{M \times N \times D} $ of image $X$, each element $f_{m,n,d}$  in the transformed features $F(X)  \in \mathbb{R}^{M \times N \times D}$ can be computed via processing each element $g_{m,n,d}$ of   $G({X}) $ by
\begin{equation}
	f_{m,n,d} = \frac{1}{1 + e^{ - g_{m,n,d} }} \ ,
\end{equation}
where $(m,n,d)$ denotes the three-dimensional index of element in input or transformed features.

\subsection{Non-local patch-based label fusion subnet}

This subnet is a deep architecture implementing non-local patch-based label fusion strategy on top of feature extraction subnet. As shown in Fig.\ref{fig:DFN_sturcture}, our NL-PLF subnet consists of {\em shift layer, distance layer, weight layer} and {\em voting layer}, and finally outputs the estimated label of target image.

\begin{figure*}[!tp]
	\subfloat[Non-local patch-based label fusion: each patch, \emph{e.g.}, patch around $q$, in a window $N_p$ around $p$ votes for the label of pixel $p$ in $T$.]{
		\label{fig:NLF-a}
		\begin{minipage}[t]{0.46\textwidth}
			\centering
			\includegraphics[width=5.8cm]{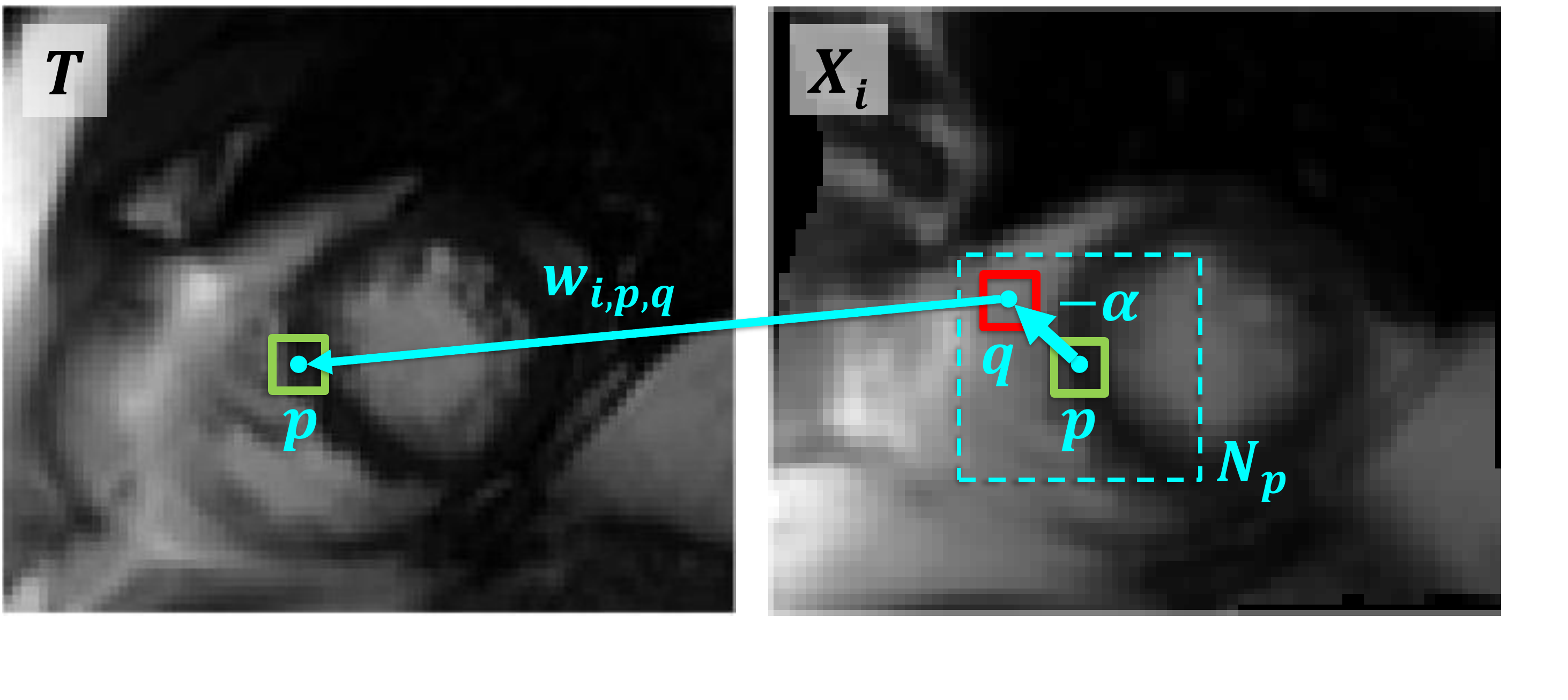}
		\end{minipage}
	}
    \hspace{2mm}
	\subfloat[Shift operation: feature distance between $p$ of $T$ and $q$ of $X_i$ in (a) equals to feature distance of $T$ and $S^{\alpha} (X_i)$ at $p$.]{
		\label{fig:NLF-b}
		\begin{minipage}[t]{0.46\textwidth}
			\centering
			\includegraphics[width=5.8cm]{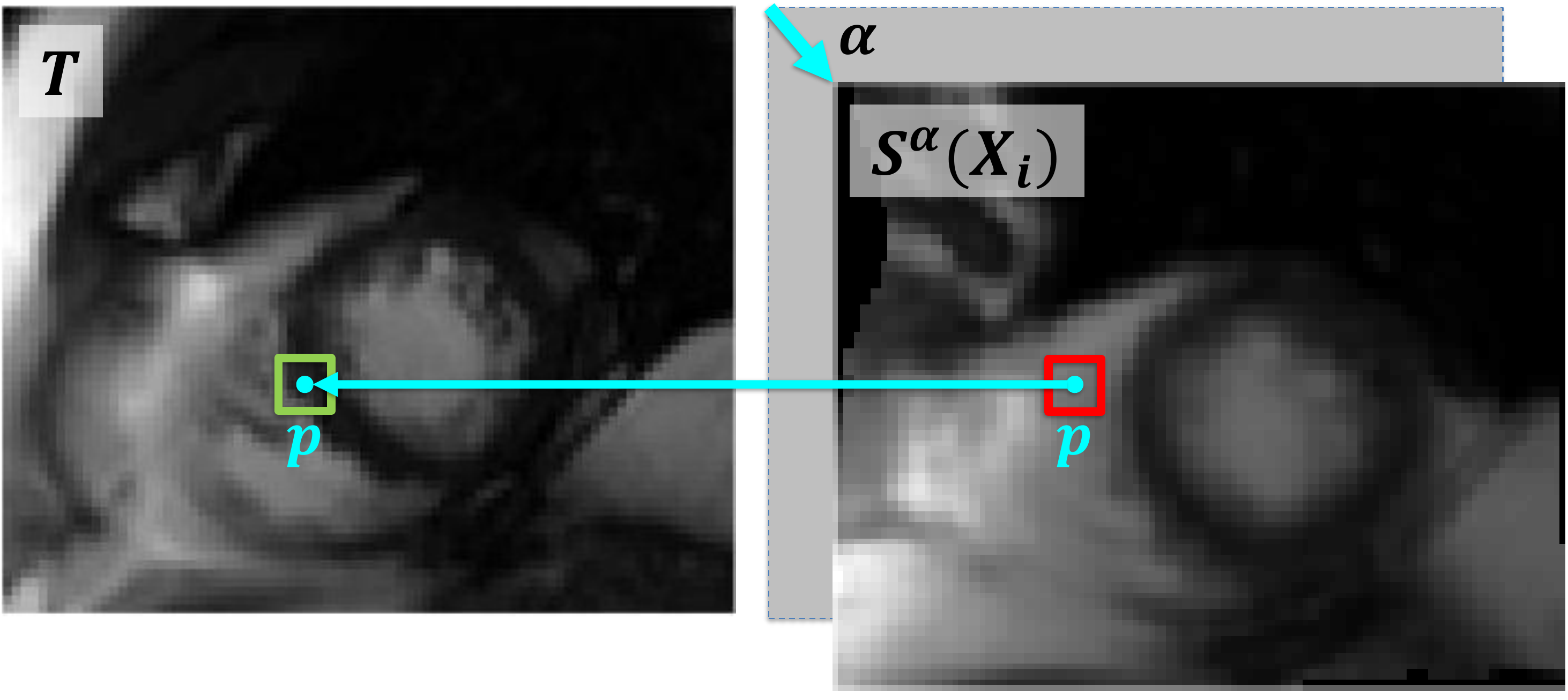}
		\end{minipage}
	}
	\caption{Illustration of non-local patch-based label fusion and shift operation. }
	\label{fig:NLF}
\end{figure*}

Figure~\ref{fig:NLF}(a) shows the idea of non-local patch-based label fusion strategy. For simplicity, we use the notation $p$ to denote a pixel's spatial position instead of $(m,n)$.
To account for atlas-to-target image registration errors, all the pixels in a search window around the pixel $p$ in warped atlas images $\{X_i\}_{i=1}^K$ are considered as the potential corresponding pixels for the pixel $p$ in target image $T$. Therefore, the labels of these pixels in atlas images are fused to produce the estimated target label at pixel $p$.

Contrary to the {hand-crafted} features adopted in \cite{Bai2015}, deep fusion net computes the fusion weights using the deep features extracted by the feature extraction subnet.
The fusion weights are defined as the normalized distances between these deep features, and the normalization is accomplished by softmax operator, enforcing that the summation of all fusion weights related to each pixel $p$ in target image $T$ is one.
More precisely, the fusion weight of pixel $q$ in warped atlas image $X_i$ for predicting the label of pixel $p$ in target image $T$ is determined as
\begin{equation} \label{eqn:wei}
	w_{i,p,q}(\Theta) = \frac{\exp(-||F_p(T; \Theta) - F_q(X_{i}; \Theta)||_2^2)}{\sum_{j} \sum_{q' \in N_p}{\exp(-||F_p( T; \Theta) - F_{q'}( X_{ j}; \Theta)||_2^2)}} \ ,
\end{equation}
where $\Theta$ is the network parameters, {\em i.e.}, filters and biases, in feature extraction subnet. $F_p( T; \Theta)$ is the extracted feature vector of image $ T$ at pixel $p$, and $N_p$ is the search window around pixel $p$.

Hence, the estimated label of pixel $p$ in target image $T$ can be written as
\begin{equation} \label{eqn:estlabel}
	\hat{L}_p( T; \Theta) = \sum_{i}\sum_{q \in N_p} w_{i, p,q} (\Theta) L_q( X_i) \ ,  
\end{equation}
where $L_q(X_i)$ is the label of atlas image $X_i$ at pixel $q$.

The objective of learning deep fusion net is to enforce that the estimated label of target image $T$ in Eqn.(\ref{eqn:estlabel}) should approximate the ground-truth target label ${L(T)}$ as close as possible.
Therefore, a {\em loss layer} for measuring the approximation error is defined as a $L_2$ loss:
\begin{equation} \label{eqn:loss}
E(\hat{{L}}(T; \Theta),{L(T)}) = \frac{1}{P} \sum_p \big\Vert \hat{{L}}_p(T;\Theta) - {L_p(T)} \big\Vert _2 ^2 \; ,
\end{equation}
where  $P$ denotes the number of pixels in target image $T$.
Our task in network training is to minimize this loss function on a training set {\em w.r.t.} the network parameters $\Theta$ using back-propagation. 
For notational convenience, we will omit network parameters $\Theta$ in the rest of paper.

Based on the formula of label prediction in Eqn.~(\ref{eqn:estlabel}), our network output $\hat{L}_p(T)$  is very often a hard 0 or 1 rather than the soft probability in $(0,1)$ of a standard soft-max classifier, and we expect the $L_2$ loss to be more computationally stable and smoother to optimize than the commonly used cross-entropy loss. Please refer to section 1 of appendix for gradient computations using $L_2$ and cross-entropy losses.
In preliminary experiments, we also tried several different loss functions, \emph{e.g.}, $L_1$, hinge, Dice and log losses, and $L_2$ loss behaved marginally better. Please refer to section~\ref{sec:comp_loss_layer} for these experiments.

To incorporate Eqn.(\ref{eqn:wei}) into the neural network as a differentiable function, we decompose the computation of the fusion weights into several successive simple operations, modeled as {\em shift layer}, {\em distance layer} and {\em weight layer}. Each operation and the gradient of its output {\em w.r.t.} input can be {easily calculated using standard deep-learning libraries}.
Figure~\ref{fig:NLF} shows our motivation for this decomposition in detail. Instead of directly computing the feature distance of pixel $p$ in target image $T$ and pixel $q$ in atlas image $X_i$, shown in Fig.~\ref{fig:NLF}(a), we can equivalently compute the per-pixel feature distance at the pixel $p$ between target image $T$ and the shifted atlas image $X_i$ by the shift vector $\alpha = p - q$, as shown in Fig.~\ref{fig:NLF}(b).
Suppose that the search window width is $2t+1$. To calculate the fusion weights using Eqn.(\ref{eqn:wei}) in deep networks, we first shift each feature map of $X_i$ by each shift vector $\alpha $ within the non-local region $ R_{nl} =  \left\lbrace (u,v) \in \mathbb{Z}^2| -t \leq u,v \leq t\right\rbrace$ using a {\em shift layer}, then compute the pixel-wise feature distances using a {\em distance layer}, and finally transform these distances to fusion weights using a {\em weight layer}.

\subsubsection{Shift layer}
\label{ShiftLayer}
\begin{figure*}[!tp]
	\setlength{\abovecaptionskip}{2mm}
	\setlength{\belowcaptionskip}{0mm}
	\centering
	\includegraphics[width=11cm]{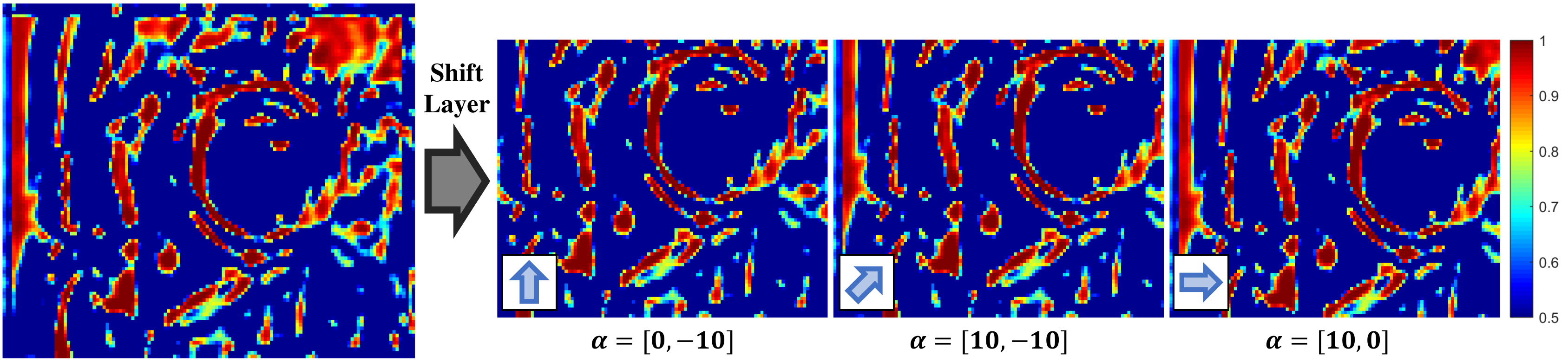}
	\caption{Illustration of shift operator $S^{\alpha}$ with three different values of $\alpha$. For one feature map (left), these shift {operators} shift it along different vectors $\alpha$ and generate shifted feature maps (right). Note that the areas going out of the border are carefully cropped.}
	\label{fig:shift_operation}
\end{figure*}

The shift layer spatially shifts the extracted features or label map of each atlas image.  Given the extracted features $F({X}_i) \in \mathbb{R}^{M \times N \times D}$ of atlas image ${X}_i$, this layer generates $(2t+1) \times (2t+1)$ spatially shifted features along each shift vector $\alpha \in R_{nl}$. Each output of the shift layer can be written as $ S^{\alpha} (F(X_i)) $, where $S^{\alpha}$ denotes the shift operator with a shift vector $\alpha$. Figure~\ref{fig:shift_operation} illustrates an example of shift operator $S^{\alpha}$ with different values of $\alpha$ on one feature map. Each shift operator $S^{\alpha}$ is actually an identity function between shifted features, and thus the gradients can be propagated back accordingly. Note that careful cropping is needed for areas going out of the border due to shift operators.

\subsubsection{Distance layer}
The distance layer computes the pixel-wise feature distance at each pixel $p$  between shifted features $S^{\alpha}(F({X}_i))$ of atlas ${X}_i$ and target's features $F({T})$ as
\begin{equation}
D^\alpha_{p} ({T}, {X}_i)= \big\Vert S^{\alpha}_p (F({X}_i)) - F_p ({T}) \big\Vert _2^2 \; ,
\end{equation}
where $S^{\alpha}_p (F({X}_i))$ denotes the feature vector of shifted features $S^{\alpha}(F({X}_i))$ at pixel $p$. It is defined as the pixel-wise $L_2$ distance, and the gradient of this layer can be simply derived \textit{w.r.t.} both $S^{\alpha}(F({X}_i))$ and $F({T})$.

\subsubsection{Weight layer}
The weight layer maps the feature distances to fusion weights using softmax operation. The fusion weight of pixel $q$ ($q = p - \alpha ,\alpha \in R_{nl})$ in atlas image $ X_i$ for predicting the label of pixel $p$ in target image $ T$ can be written as
\begin{equation}
w_{i, p, q} = w^{\alpha}_p( X_i)= \frac{e^{-D^{\alpha}_{p} ( T,  X_i)}}{\sum_{j}  \sum_{\alpha ' \in R_{nl}}e^{-D^{\alpha '}_{p} ( T,  X_j)}} \; .
\end{equation} 
This softmax operation is a common layer in architecture of deep networks, and the gradient of this layer \textit{w.r.t.} the input $D^{\alpha}_{p} ( T,  X_k)$ can be easily derived \citep{Goodfellow2016}, where $k$ is any atlas index.

\subsubsection{Voting layer}
The voting layer estimates the label of target image $T$ at pixel $p$ by
\begin{equation}
\hat{{L}}_p(T)= \sum_i \sum_{\alpha \in R_{nl}} w_p^\alpha({X}_i)  S_p^\alpha({L(X_i)}) \; .
\end{equation}
As a linear operation, the gradient of this layer can be easily derived \textit{w.r.t.} the input $w_p^\alpha({X}_i)$.

\textbf{\textit{Summary:}}
The non-local patch-based label fusion subnet successively processes the extracted features and warped atlas labels by \emph{shift}, \emph{distance}, and \emph{weight layers} to output fusion weights, which are then utilized by \emph{voting layer} to estimate the target label. This subnet implements Eqn.(\ref{eqn:wei}) by using the above simple layers, and the gradients of these layers can be {easily calculated using standard deep-learning libraries}.

\subsection{Deep fusion net for binary and multi-class segmentation} 
The deep fusion net can be adapted to both tasks of binary and multi-class segmentation.  This can be uniformly achieved by representing the segmentation label $L_p$ of target or warped atlas pixel $p$ using probability vector in the voting layer. Suppose that there are $C$ classes for segmentation, the segmentation label of pixel $p$ can be represented by  $L_p = (a_1, \cdots, a_C)^\top$ where $a_i \geq 0$ and $\sum_i a_i = 1$.  $a_i$ represents the probability of pixel $p$ belonging to class $i$. 
For binary segmentation, $L_p = (a, 1-a)^{\top}$ where $a$ represents the probability of pixel $p$ belonging to object of interest.

\subsection{Network training}
\label{network_training}
We learn the network parameters $\Theta$ by minimizing the loss in Eqn.(\ref{eqn:loss})  {\em w.r.t.} $\Theta$ using back-propagation.
Given a number of atlases for training, each atlas is selected as the target image in turn, and then the remaining atlases are registered to this target image as the warped atlases.
Suppose that $i$-th atlas is picked as the target image $T$ with ground-truth label $ L(T)$, the corresponding warped atlases are denoted by $\mathcal A_i = \left\lbrace {X}_{j}, {L}({X_j})|  j=1,2,...,K, j \neq i \right\rbrace $, where $K$ is the total number of atlases.

We use stochastic gradient descent in training, and each triplet of $(\mathcal{A}_{i}, T,{L}(T)  )$ is taken as a batch. 
Instead of using all warped atlases within $\mathcal A_i$ in each batch, which may require a large GPU memory and long training time, we randomly sample $K_0$ ($K_0=5$) warped atlases for training,  according to a distribution proportional to the normalized mutual information between warped atlas images $\left\lbrace {X}_{j} |  j=1,2,...,K, j \neq i \right\rbrace $ and target image $T$. 
This random atlas selection strategy enriches the diversity of warped atlases for target image at training phase. Since there may be discrepancy between distributions of training and testing data, this strategy may improve the generalization ability of deep fusion net for testing data. We will evaluate the atlas selection strategy at training phase in section~\ref{sec:impact_atlas_selec}. Many randomized techniques were also proposed in deep learning to improve network generalization ability, \emph{e.g.}, dropout~\citep{Srivastava2014}, data augmentation~\citep{Vincent2008}, \emph{etc}.

\subsection{Network testing}
At testing phase, the learned deep fusion net loads a test sample (a target image and its warped atlases) and outputs the estimated target label. Similar to multiple atlas selection approaches~\citep{Sanroma2014,Albert2013,Aljabar2009}, we only pick a few most similar atlases for the target image in label fusion, which is an atlas selection problem. 
As discussed in section~\ref{sec:framework}, feature extraction subnet actually learns an embedding space at training process, in which pixel-wise $L_2$ feature distance within distance layer corresponds to image similarity. We can naturally define a deep feature distance between a target image $T$ and its warped atlas image $X_i$ as
\begin{equation}
d_F( T,  X_i) = \big\Vert F( T) - F( X_i) \big\Vert ^2_F \ ,
\end{equation}
where $F(\cdot)$ is the extracted features using well-trained feature extraction subnet at training process. At testing phase, we take the top-$k$ atlases with smallest deep feature distances as the selected atlases for a target image, which are then fed into the learned deep fusion net to estimate the target label.
Note that the number $k$ of selected atlases at testing process does not need to be same as $K_0$ used at training phase. 
In fact, one could intuitively expect that $K_0$ at training phase would make the features to be learned in a way that the network operates best at testing phase with $k = K_0$. However, we show empirically in section~\ref{sec:impact_atlas_selec} that a larger number $k$ at testing phase would generally produce better segmentation accuracies. This happens possibly because of the random atlas selection strategy utilized at training phase, which largely extends the diversity of warped atlases for target image at training phase and enables the network to utilize more warped atlases at testing process.

\section{Implementation details and computational requirements}
Our current implementation is based on the MatConvNet library~\footnote{http://www.vlfeat.org/matconvnet/}, and all experiments are performed on a Dell Precision T7910 workstation with GeForce GTX TITAN X (12GB) on an Ubuntu platform.
In the current study, our proposed method processes cardiac MR images slice by slice, and the GPU memory requirement for processing one target image is about $O \big(MND{K_0} {(2t+1)^2} \big)$. 
The averaged size of target images for numerical experiments in section~\ref{sec:experiments} is about $110\times140$ for SATA-13 dataset and $135\times155$ for LV-09 dataset.
In practice, for a target image in size of $120 \times 140$ with $K_0 = 5$, a deep fusion net with network parameters in section~\ref{sec:para_settings} requires about 4GB GPU memory for one forward and backward processes, which respectively {take} about 0.30 second and 0.31 second.
In this case, the maximum allowed number of selected atlases $K_0$ for training is up to 42 in our computational platform.

\section{Experiments}
\label{sec:experiments}
In this section, we evaluate the performance of deep fusion net on two cardiac MR datasets for left ventricle (LV) segmentation. 
In the following paragraphs, we will compare our method with other segmentation methods and further investigate the performance of our deep fusion net with variants of architecture, loss, atlas selection strategy, cross-dataset evaluation on these two datasets.
In appendix, we also provide additional experiments, including justification of the effectiveness of learned deep features as similarity measure, and all the $p\mbox{-values}$ {appearing} in the following paragraphs.

\subsection{Datasets}
\label{dataset}
\subsubsection{SATA-13 dataset}
The MICCAI 2013 SATA Segmentation Challenge (SATA-13) provides a cardiac dataset for LV segmentation. The samples in this dataset are randomly selected from DETERMINE (Defibrillators to Reduce Risk by Magnetic Resonance Imaging Evaluation) in Cardiac Atlas Project (CAP).
The cardiac MR images are acquired using the steady-state free precession (SSFP) pulse sequence, and each image is acquired during a breath-hold of 8-15 seconds duration. Sufficient short-axis slices are obtained to cover the whole heart, and the MR parameters vary between cases. 
Typically, each cine image sequence has about 25 frames, and each frame has about 10 slices. The size of slice ranges from $138\times192$ to $512\times512$. The slice thickness is less than 10 mm, and the gap between slices is less than 2 mm.

The SATA-13 dataset includes 83 training subjects and 72 testing subjects. 
The ground-truth myocardium segmentation masks for all frames in training subjects are provided, while we evaluate our deep fusion net only on end-diastole (ED) frame, as a common practice in the literature~\citep{Petitjean2011,Shi2014}. The experiment is performed over 83 training subjects using 5-fold cross validation, \emph{i.e.}, one fifth of training subjects are taken as validation set and the remaining four fifths of subjects are taken for learning deep fusion net in each fold. 
The averaged 3D Dice metric (ADM) and averaged 3D Hausdorff distance (AHD) over the validation sets in five folds are taken as the final accuracies. {For one subject in validation sets}, assume that $\Omega_{gt}$ and $\Omega_{et}$ are respectively its ground-truth and estimated segmentations represented by sets of pixels labeled as object of interest, the {Dice metric (DM)} and {Hausdorff distance (HD)} are defined as $DM(\Omega_{gt},\Omega_{et})= \frac{2 | \Omega_{gt} \cap \Omega_{et} | }{|\Omega_{gt}|+|\Omega_{et}|}$ and $HD(\Omega_{gt},\Omega_{et})= \max \big( \max\limits_{p\in \Omega_{gt}} ( \min\limits_{q\in \Omega_{et}}  d(p,q) ), \max\limits_{q\in \Omega_{et}} ( \min\limits_{p\in \Omega_{gt}}  d(p,q) ) \big)$, where $|\cdot|$ denotes the number of elements in a set and $d(p,q)$ denotes the Euclidean distance between coordinates of pixels $p$ and $q$. The Hausdorff distance is computed in spatial resolution of millimeter obtained from the DICOM file.

\subsubsection{LV-09 dataset}
The MICCAI 2009 LV Segmentation Challenge (LV-09) dataset~\citep{Radau2009}  is provided by Sunnybrook Health Sciences Center, and contains 45 subjects with expert annotations, respectively in subsets of ``training'', ``testing'' and ``online''. The cardiac cine-MR short-axis images are acquired using SSFP pulse sequence with a 1.5T GE Signa MRI. All images are obtained during 10-15 seconds {of breath holding} with a temporal resolution of 20 cardiac phases over the heart cycle. Each subject contains all slices at end-diastole (ED) and end-systole (ES) frames, and each frame contains 6-12 {short-axis} images obtained from the atrioventricular ring to the apex (thickness = 8 mm, gap = 8 mm, FOV = $320 \times 320 $ mm, matrix = $ 256 \times 256$).
Both endocardial and epicardial contours are drawn by experienced cardiologists in all slices at ED frame, while only endocardial contours are given at ES frame.

In the experiment, as in~\cite{Avendi2016}, we utilize 15 training subjects for learning deep fusion net, and 30 subjects in testing and online sets for evaluating the performance.
The standard evaluation scheme of MICCAI 2009 LV Segmentation Challenge~\citep{Radau2009} is  utilized in our comparison, which is based on the following three measures: 1) percentage of ``good" contours, 2) averaged Dice metric (ADM) of the ``good" contours, and 3) averaged perpendicular distance (APD) of the ``good" contours. A contour is classified as ``good" if APD is less than 5 mm.
The Dice metric and perpendicular distance {are} calculated for each 2D slice separately, and the evaluation measures (\emph{i.e.}, ``{Good}" percentage, ADM and APD) are averaged over all slices within ED and ES frames of all subjects in testing or online sets.

\subsection{Preprocessing}
\label{Preprocessing}
Before starting to learn deep fusion net, atlas images are registered to the target image using ITK software package~\footnote{http://www.itk.org/}, and we utilize two different registration frameworks to test the robustness of our proposed method.

\textit{Landmark-based (LB) registration:} Each atlas subject is warped to the target subject using landmark-based registration with 3D affine transformation, and five landmarks are manually labeled at both atlas and target images, as in~\cite{Bai2013}. These landmarks are also used to crop the {region of interest (ROI)} from complex backgrounds to reduce the computational cost of registration.
To compensate for the potential inter-slice shift, 2D B-spline registration is then utilized on each pair of slices  in atlas and target subjects respectively.

\textit{Landmark-free (LF) registration:} The ROI is cropped by a bounding box on each subject determined by two corner points, and then each atlas subject is warped to the target subject by 3D affine registration without using any landmarks. Normalized mutual information (NMI) is taken as the similarity metric. At last, 2D affine registration and 2D B-spline registration are successively applied to each slice for reducing the potential inter-slice shift.

Generally speaking, LB {registration} framework would produce more accurate results due to the guidance of landmarks. In each process of registration, with the estimated motion between atlas and target images, each atlas label is  warped to provide an estimate for the target label using the {corresponding} motion field. All the warped estimates {for target label} are fused by deep fusion net to generate a final prediction of {the} target label.

\subsection{Parameter settings}
\label{sec:para_settings}
Empirically, we fix the learning rate of deep fusion net to $5 \times 10^{-7}$. Unless otherwise stated, the numbers of selected atlases at training and testing phases, \emph{i.e.}, $K_0$ and $k$, are respectively 5 and 10, and the size of search window in the shift layer is $7 \times 7$.
As for the feature extraction subnet, we utilize 4 convolutional layers with strides of 1, and these layers respectively have 64 filters in size of $5 \times 5 \times 1$, 64 filters in size of $5 \times 5 \times 64$, 128 filters in size of $5 \times 5 \times 64$, and 128 filters in size of $5 \times 5 \times 128$.

\subsection{Segmentation accuracy}

\subsubsection{SATA-13 dataset}
\label{sec:SATA13_accuracy}
For SATA-13 dataset, we compare the segmentation accuracy of deep fusion net with the results of majority voting (MV), patch-based segmentation (PB)~\citep{Coupe2011}, as well as SVM-based segmentation with augmented features (SVMAF)~\citep{Bai2015}.
Another version of deep fusion net which uses normalized mutual information (NMI) as similarity measure for atlas selection at testing process, denoted as DFN\_NMI, is also included in the comparison.
The results of MV, PB and SVMAF are reproduced by the published codes~\footnote{\url{http://wp.doc.ic.ac.uk/wbai/software/}} using the same parameter settings in \cite{Bai2015}.
The segmentation accuracy is evaluated by averaged Dice metric (higher value means better performance) and averaged Hausdorff distance (lower value means better performance) of the myocardium across ED frame and all the testing subjects in 5-fold cross validation.
A paired-samples t-test is conducted to compare the performance of segmentation approaches.

\begin{table*}[!htbp]\footnotesize
	\centering
	\renewcommand\arraystretch{1.4}
	\begin{tabu}{l*{3}{@{\hspace{1.2em}} c}}
		\tabucline[1.5pt]{-}
		Method       & \tabincell{c}{Registration \\ framework}     & \tabincell{c}{Averaged \\ Dice metric}   & \tabincell{c}{Averaged \\ Hausdorff distance} \\
		\tabucline[1.2pt]{-}
		CNN &  -         &  0.695(0.092)         &  52.33(11.59)              \\
		MV & LB       &  0.747(0.057)         &  12.16(3.87)               \\
		PB & LB       &  0.761(0.056)         &  12.11(3.86)               \\
		SVMAF & LB    &  0.777(0.057)         &  17.31(4.40)               \\
	\hdashline
DFN &LB      & \textbf{0.833}(\textbf{0.039}) &  \textbf{11.35}(\textbf{3.53})            \\ 
		DFN &LF      &  0.815(0.044)         &  18.33(8.98)               \\ 
		DFN\_NMI &LB &  0.831(0.039)         &  11.98(3.84)               \\
		DFN\_NMI &LF &  0.802(0.051)         &  27.00(12.75)              \\

		\hdashline
		DFN(crossDS) & LB &  0.815(0.040)         &  11.76(3.70)               \\
		DFN\_NMI(crossDS) & LB & 0.809(0.040)     &  12.46(4.19)               \\
		\tabucline[1.5pt]{-}
	\end{tabu}
	\caption{{{Myocardium}} segmentation accuracies of different methods on SATA-13 dataset. Each cell is formatted as ``mean (standard deviation)". ``LB", ``LF" and ``-" respectively denote landmark-based registration, landmark-free registration and registration-free method.}
	\label{tab:SATA13_acc}
\end{table*}

Table~\ref{tab:SATA13_acc} reports the quantitative segmentation accuracies of different methods using LB or LF registrations introduced in section~\ref{Preprocessing}.
The experimental results show that our DFN achieves better performance than the  methods of MV, PB and SVMAF in both ADM~{($p\mbox{-value} < 0.001$)} and AHD~{($p\mbox{-value} < 0.05$)} when using LB registration. 
In addition, our DFN and DFN\_NMI using LF registration obtain higher accuracies in ADM than these compared methods of SVMAF, PB, MV, CNN using LB registration~{($p\mbox{-value} < 0.001$)}.
{As for similarity measures in atlas selection, by comparing DFN (using deep features for atlas selection) with DFN\_NMI (using NMI for atlas selection), our defined deep feature-based atlas selection works significantly better in terms of both two metrics~{($p\mbox{-value} < 0.001$)} when using LF registration, improving ADM from 0.802 to 0.815 and AHD from 27.00 to 18.33. 
However, this improvement when using landmark-based (LB) registration is marginal by comparing DFN and DFN\_NMI with LB registration, possibly because LB  registration achieves better registration quality  decreasing the dependency of segmentation accuracy on a better atlas selection strategy.}

Convolutional neural network (CNN) is also included in this comparison to evaluate the effect of NL-PLF subnet, and we restrict that the CNN and our DFN have identical network capacity of feature extraction, \emph{i.e.}, the compared CNN in Table~\ref{tab:SATA13_acc} has the same feature extraction subnet as ours in section~\ref{sec:para_settings}, which is then followed by a convolutional layer and a softmax layer to output the target label.
The meta-parameters of CNN, \emph{e.g.}, learning rate, are reconfigured to the best of our abilities.
Experimental results show that our DFN performs significantly better than CNN in all metrics~{($p\mbox{-value} < 0.001$)}, indicating the effectiveness of NL-PLF subnet.

\begin{figure*}[!tp]
	\setlength{\abovecaptionskip}{1mm}
	\setlength{\belowcaptionskip}{0mm}
	\centering
	\subfloat[Averaged Dice metric]{
		\label{fig:SATA13_box_DR}
		\begin{minipage}[t]{0.48\textwidth}
			\centering
			\includegraphics[width=6.0cm]{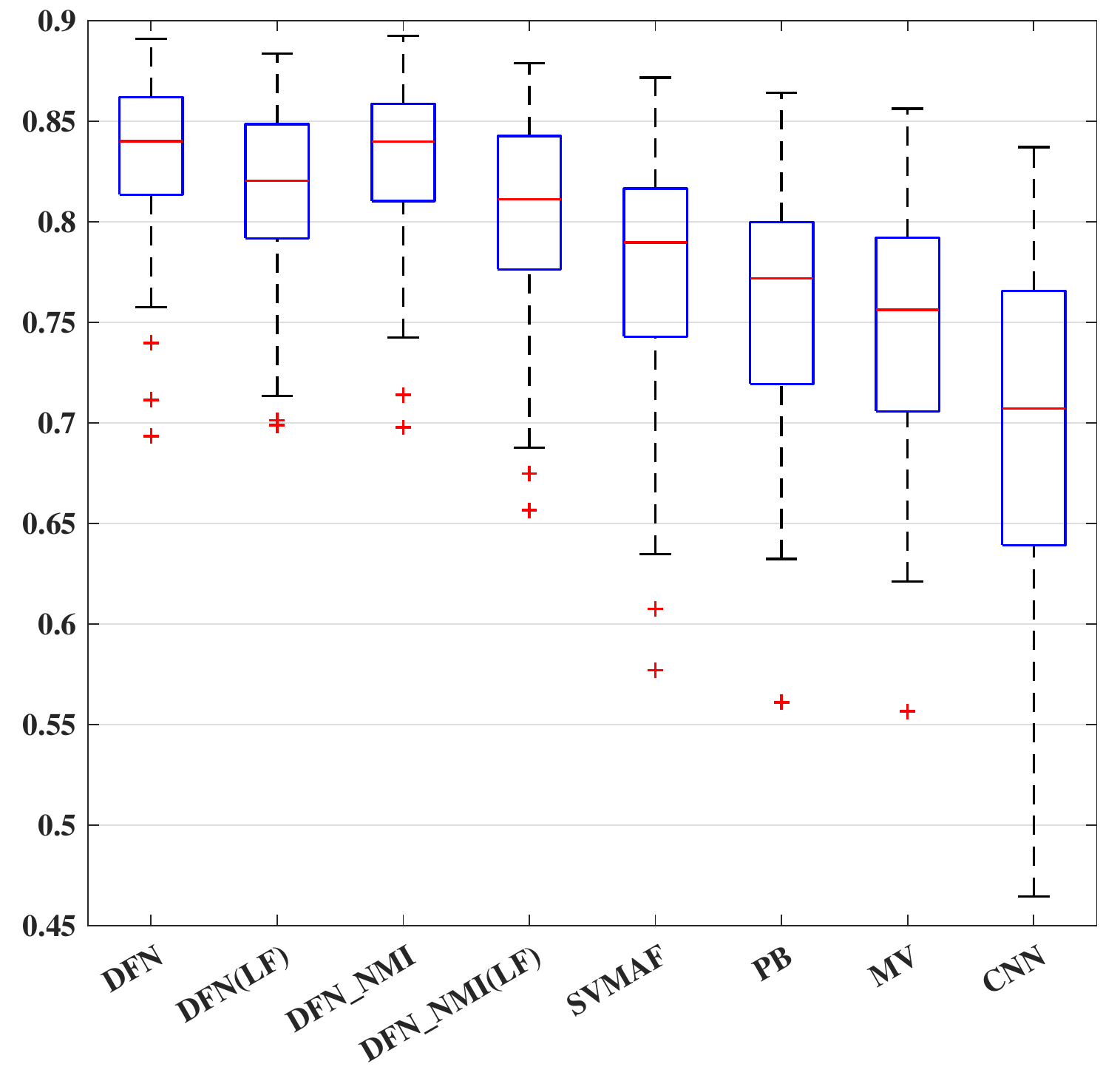}
		\end{minipage}
	}
	\subfloat[Averaged Hausdorff distance]{
		\label{fig:SATA13_box_HD}
		\begin{minipage}[t]{0.48\textwidth}
			\centering
			\includegraphics[width=6.0cm]{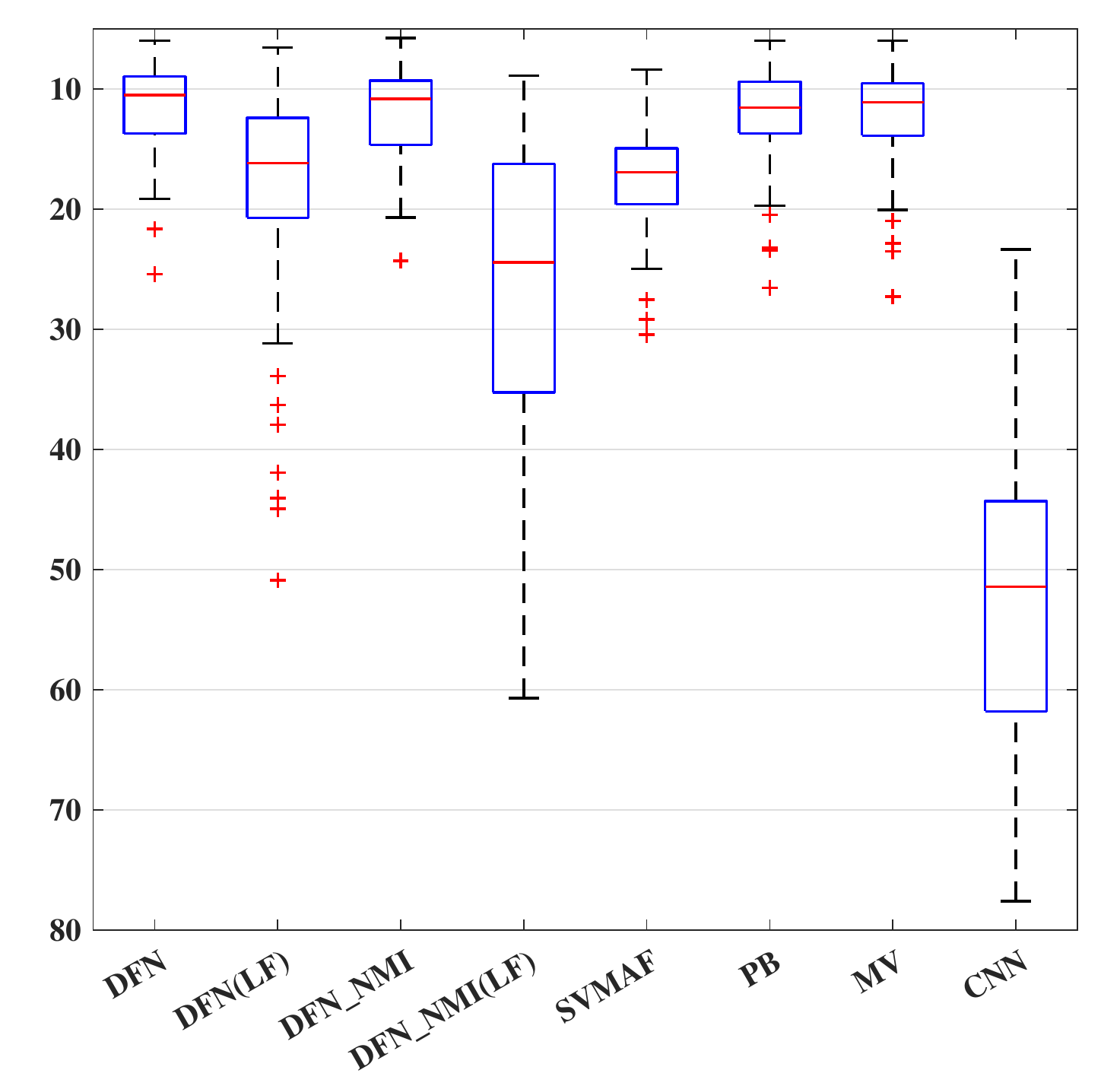}
		\end{minipage}
	}
	\caption{Comparison of myocardium segmentation accuracies for different methods on SATA-13 dataset in terms of (a) ADM and (b) AHD. The red central mark of the box is the median and the edges are the 25th and 75th percentiles, while the red plus signs represent outliers.}
	\label{fig:SATA13_box}
\end{figure*}

\begin{figure*}[!bp]
	\setlength{\abovecaptionskip}{1mm}
	\setlength{\belowcaptionskip}{0mm}
	\centering
	\includegraphics[width=12.2cm]{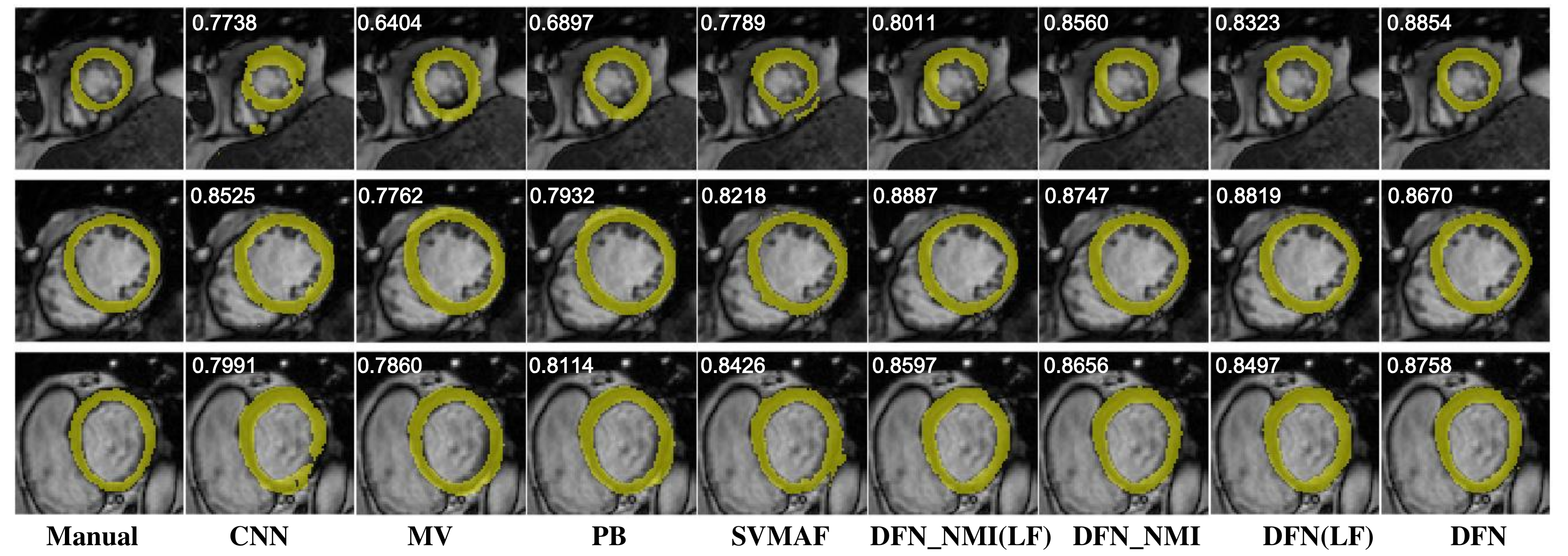}
	\caption{Visual comparison of {myocardium} segmentation results using different methods on SATA-13 dataset. Please refer to the electronic version for better comparison. Top to bottom: the basal, mid-ventricular and apical slices from the same testing subject. Left to right: manual labeling, convolutional neural network (CNN), majority voting (MV), patch-based method (PB), SVM segmentation with augmented features (SVMAF), DFN\_NMI using LF registration, DFN\_NMI, DFN using LF registration, and DFN. The {small white text} on each sub-image is the corresponding 2D Dice metric {value}.}
	\label{fig:SATA13_visual}
\end{figure*}

Figure~\ref{fig:SATA13_box} compares the segmentation accuracies in the box-plot, showing that our DFN based on LB registration achieves significantly higher median values than the compared methods in ADM, and is marginally better in AHD compared with PB and MV.
Figure~\ref{fig:SATA13_visual} shows visual examples of the segmentation results by different methods on the basal, mid-ventricular and apical slices from a testing subject.

Furthermore, to compare with other state-of-the-art methods, we trained a deep fusion net on Cardiac Atlas Project (CAP) training set of MICCAI 2013 SATA Segmentation Challenge and tested its performance on CAP testing set of the challenge, whose ground-truth segmentations are unknown to challenge participants. We submitted our segmentation results to the challenge website, and the corresponding testing accuracies were evaluated by the website and published on its leaderboard~\footnote{Old website: http://masi.vuse.vanderbilt.edu/submission/leaderboard.html}$^{,}$\footnote{New website: https://www.synapse.org/\#!Synapse:syn3193805/wiki/217788 (Last accessed: {10 July 2018})}.
Our proposed deep fusion net based on LB registration (referred as DeepMAS\_LB) achieved 0.815 in Dice metric,  ranking first among the submitted results at the time of writing.

\begin{table*}[!bp]
	\centering
	\renewcommand\arraystretch{1.4}
	\resizebox{\textwidth}{!}{
		\begin{tabu}{l @{\hspace{0.4em}} c @{\hspace{0.4em}} c  @{\hspace{0.3em}} c  @{\hspace{0.8em}} c @{\hspace{0.3em}} c  @{\hspace{0.8em}} c  @{\hspace{0.3em}} c}
			\tabucline[1.5pt]{-}
			\multirow{2}{*}{Method}
			& Registration
			& \multicolumn{2}{c}{``Good" percentage} \hspace{0.4em}
			& \multicolumn{2}{c}{Epicardium ADM} \hspace{0.4em}
			& \multicolumn{2}{c}{Epicardium APD} \hspace{0.4em}     \\
			& Framework
			&  Testing					 &  Online 
			&  Testing					 &  Online 
			&  Testing					 &  Online                  \\
			\tabucline[1.2pt]{-}
			MV          &  LB
			&  94.60(8.17)               &  92.34(13.16)
			&  0.93(0.02)                &  0.92(0.02)
			&  2.57(0.41)                &  2.80(0.42)              \\
			PB          &  LB
			&  94.60(8.17)               &  92.34(13.16)
			&  0.93(0.02)                &  0.92(0.02)
			&  2.54(0.41)                &  2.77(0.43)              \\
			SVMAF       &  LB
			&  95.43(8.03)               &  84.84(21.74)
			&  0.93(0.01)                &  0.93(0.01)
			&  2.85(0.50)                &  3.01(0.47)              \\
			DBNLS       &  -
			&  94.65(6.18)               &  84.32(23.45)
			&  0.93(0.02)                &  0.93(0.03)
			&  2.08(0.60)                &  2.05(0.61)              \\
			{DBNLS(semiauto)} &  -
			& \textbf{100}(\textbf{0})   &  \textbf{100}(\textbf{0})
			&  0.94(0.01)                &  0.94(0.02)
			&\textbf{1.73}(\textbf{0.28})&\textbf{1.90}(\textbf{0.53})\\
			\hdashline
			DFN         &  LB
			& \textbf{100}(\textbf{0})   &  93.62(13.55)
			&  0.95(0.01)                & \textbf{0.95}(\textbf{0.01})
			&  1.90(0.22)                &  1.98(0.27)              \\
			DFN         &  LF
			&  78.90(20.58)              &  65.40(29.46)
			&  0.94(0.01)                &  0.93(0.04)
			&  2.55(0.80)                &  2.73(0.68)              \\
			DFN(multi)  &  LB
			&  97.43(7.64)               &  96.26(8.19)
			&  0.95(0.01)                &  \textbf{0.95}(\textbf{0.01})
			&  2.11(0.26)                &  2.15(0.46)              \\
			DFN\_NMI    &  LB
			&  98.67(3.52)               &  92.53(13.58)
			&\textbf{0.96}(\textbf{0.01})& \textbf{0.95}(\textbf{0.01})
			&  1.88(0.18)                &  2.01(0.32)              \\
			DFN\_NMI    &  LF
			&  66.15(21.00)              &  50.54(27.29)
			&  0.93(0.03)                &  0.93(0.03)
			&  2.80(0.86)                &  2.91(0.79)              \\
			DFN\_NMI(multi)  &  LB
			&  97.43(7.64)               &   95.02(11.91)
			&  0.95(0.01)                &   0.94(0.01)
			&  2.15(0.27)                &   2.17(0.41)             \\
			\tabucline[1.5pt]{-}
	\end{tabu}}
	\caption{{Epicardium} segmentation accuracies of different methods on {testing and online sets} of LV-09 dataset, where only training set is used as training data. Each cell is formatted as ``mean (standard deviation)".}
	\label{tab:LV09_epi_acc}
\end{table*}

\subsubsection{LV-09 dataset}
\label{sec:LV09_accuracy}
For LV-09 dataset, we respectively compare the epicardium and endocardium segmentation accuracies of the testing and online sets in Tables~\ref{tab:LV09_epi_acc} and \ref{tab:LV09_endo_acc}. 
Similar to~\cite{Ngo2017}, both ED and ES frames are evaluated to measure accuracies for endocardium segmentation, while only ED frame is evaluated for the epicardium segmentation.
Besides the traditional multi-atlas segmentation methods, we also compare two recent deep learning-based methods that achieve state-of-the-art results on this dataset, \emph{i.e.} the combined stacked autoencoder and level set method (SAELS)~\citep{Avendi2016} and the combined deep belief network and level set method (DBNLS)~\citep{Ngo2017}.
Both SAELS and DBNLS first utilize deep network to estimate an initial segmentation result, which is then refined by level set approach to produce a final estimation. More exactly, DBNLS respectively trains four DBNs for epicardium or endocardium at ED or ES frame, and SAELS respectively trains two networks for large-contour or small-contour images on endocardium segmentation.
In our experiments, we try two different versions of network training{:} first, three different deep fusion nets {are respectively trained} for epicardium at ED, endocardium at ED and endocardium at ES; second, one multi-class deep fusion net (indicated by ``multi" in tables) {is trained} for epicardium and endocardium at ED, and one deep fusion net {is trained} for endocardium at ES.
Notice that DBNLS respectively utilizes training set for model training and online set for model selection, and reports the performance on testing set, while SAELS has the same experimental setting with us, \emph{i.e.}, training on training set and reporting results on testing and online sets.
Since SAELS has not been evaluated on epicardium in \cite{Avendi2016}, we only compare its accuracy on endocardium segmentation in Table~\ref{tab:LV09_endo_acc}.

As shown in Tables~\ref{tab:LV09_epi_acc} and \ref{tab:LV09_endo_acc}, our DFN performs better than the other registration-based methods (\emph{i.e.}, MV, PB and SVMAF) on epicardium and endocardium segmentations in all metrics ($p\mbox{-value} < 0.001$) when using LB registration.
In comparison with the deep learning-based methods, the only method that outperforms our DFN is DBNLS but only when it utilizes the strong manual prior (\emph{i.e.}, manually-labeled segmentation masks as prior, referred as {DBNLS(semiauto)}). In fact, compared to DBNLS without using the manual prior (referred as DBNLS), our DFN based on LB registration achieves much higher accuracies in all metrics.
As for SAELS, our DFN using LB registration achieves comparable results on endocardium segmentation. However, SAELS utilizes post-processing to refine the initial segmentation of stacked autoencoder  by level set method.
Without post-processing, SAELS reports the ADM scores of 0.90 in testing set and 0.89 in online set, compared with 0.92 and 0.92 of ours purely based on thresholding the estimated label probability maps of deep fusion nets, as shown in Table~\ref{tab:LV09_endo_acc}.
Compared with training separate DFN (referred as DFN), the multi-class DFN produces worse APD scores for epicardium ($p\mbox{-value} < 0.001$) and is comparable in other {metrics}.

\begin{table*}[!tp]\footnotesize
	\centering
	\renewcommand\arraystretch{1.4}
	\resizebox{\textwidth}{!}{
		\begin{tabu}{l @{\hspace{0.4em}} c @{\hspace{0.4em}} c  @{\hspace{0.3em}} c  @{\hspace{0.8em}} c @{\hspace{0.3em}} c  @{\hspace{0.8em}} c  @{\hspace{0.3em}} c}
			\tabucline[1.5pt]{-}
			\multirow{2}{*}{Method}
			& Registration
			& \multicolumn{2}{c}{``Good" percentage} \hspace{0.5em}
			& \multicolumn{2}{c}{Endocardium ADM} \hspace{0.5em}
			& \multicolumn{2}{c}{Endocardium APD} \hspace{0.5em}      \\
			&  Framework
			&  Testing					 &  Online 
			&  Testing					 &  Online 
			&  Testing					 &  Online                    \\
			\tabucline[1.2pt]{-}
			MV          &  LB
			&  91.08(15.11)              &  90.32(14.64)
			&  0.87(0.05)                &  0.88(0.05)
			&  2.77(0.53)                &  2.73(0.56)                \\
			PB          &  LB
			&  92.11(12.00)              &  91.71(14.20)
			&  0.87(0.06)                &  0.88(0.05)
			&  2.76(0.54)                &  2.65(0.55)                \\
			SVMAF       &  LB
			&  94.79(11.06)              &  92.86(12.65)
			&  0.88(0.04)                &  0.89(0.05)
			&  3.00(0.51)                &  2.82(0.60)                \\
			DBNLS       &  -
			&  95.91(5.28)               &  90.54(14.40)
			&  0.88(0.03)                &  0.89(0.03)
			&  2.34(0.46)                &  2.17(0.46)                \\
			{DBNLS(semiauto)} &  -
			& \textbf{100}(\textbf{0})   & \textbf{100}(\textbf{0})
			&  0.91(0.03)                &  0.91(0.03)
			&  1.79(0.36)                &\textbf{1.78}(\textbf{0.49})\\
			SAELS       &  -
			&  97.8(4.7)                 &  95.58(6.7)
			&\textbf{0.94}(\textbf{0.02})&\textbf{0.93}(\textbf{0.02})
			&\textbf{1.7}(\textbf{0.37}) &  1.92(0.51)                \\
			SAELS(init) &  -
			&  90(10)                    &  87(12)
			&  0.90(0.1)                 &  0.89(0.03)
			&  2.84(0.29)                &  2.95(0.54)                \\
			\hdashline
			DFN         &  LB
			&  97.58(7.12)               &  96.36(8.41)
			&  0.92(0.04)                &  0.92(0.03)
			&  2.04(0.44)                &  2.06(0.58)                \\
			DFN         &  LF
			&  73.56(29.00)              &  60.87(26.48)
			&  0.91(0.06)                & \textbf{0.93}(\textbf{0.02})
			&  2.31(0.40)                &  2.41(0.47)                \\
			DFN(multi)  &  LB
			&  96.96(8.48)               & 95.70(9.12)
			&  0.92(0.04)                & 0.92(0.03)
			&  2.01(0.42)                & 2.04(0.58)                 \\
			DFN\_NMI    &  LB
			&  97.25(7.24)               &  95.41(9.43)
			&  0.92(0.04)                &  0.92(0.04)
			&  2.05(0.42)                &  2.04(0.53)                \\
			DFN\_NMI    &  LF
			&  66.61(29.59)              &  53.81(25.88)
			&  0.92(0.03)                &  0.90(0.07)
			&  2.35(0.47)                &  2.66(0.65)                \\
			DFN\_NMI(multi)  &  LB
			&  97.44(7.32)               &  94.99(10.05)
			&  0.92(0.04)                &  0.92(0.04)
			&  2.06(0.41)                &  2.02(0.50)                \\
			\tabucline[1.5pt]{-}
	\end{tabu}}
	\caption{{Endocardium} segmentation accuracies of different methods on {testing and online sets} of LV-09 dataset, where only training set is used as training data. Each cell is formatted as ``mean (standard deviation)".}
	\label{tab:LV09_endo_acc}
\end{table*}

\begin{figure*}[!hbtp]
	\setlength{\abovecaptionskip}{1mm}
	\setlength{\belowcaptionskip}{0mm}
	\centering
	\subfloat[Epicardium]{
		\label{fig:LV09_box_epi}
		\begin{minipage}[t]{0.48\textwidth}
			\centering
			\includegraphics[width=6.0cm]{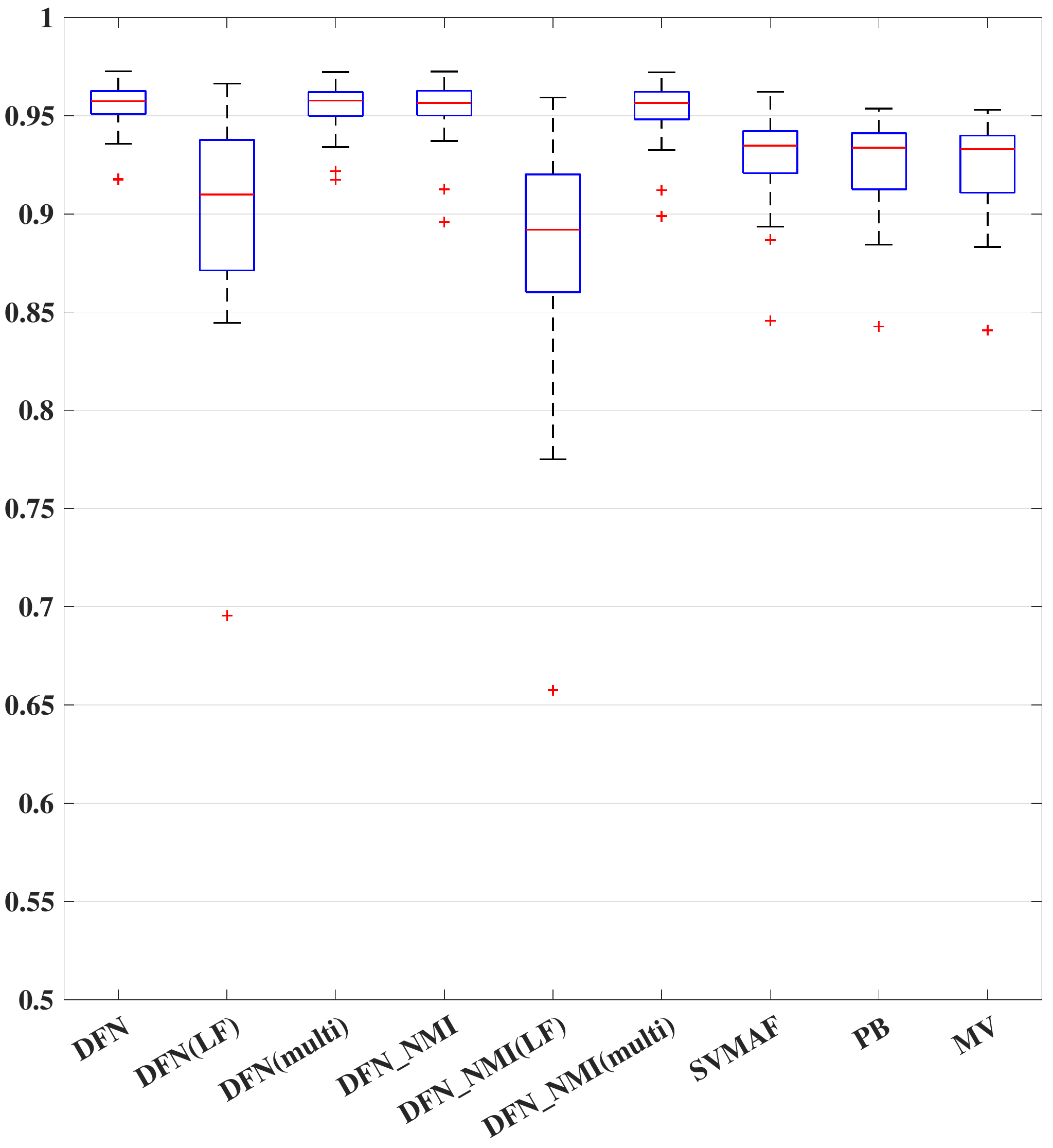}
		\end{minipage}
	}
	\subfloat[Endocardium]{
		\label{fig:LV09_box_endo}
		\begin{minipage}[t]{0.48\textwidth}
			\centering
			\includegraphics[width=6.0cm]{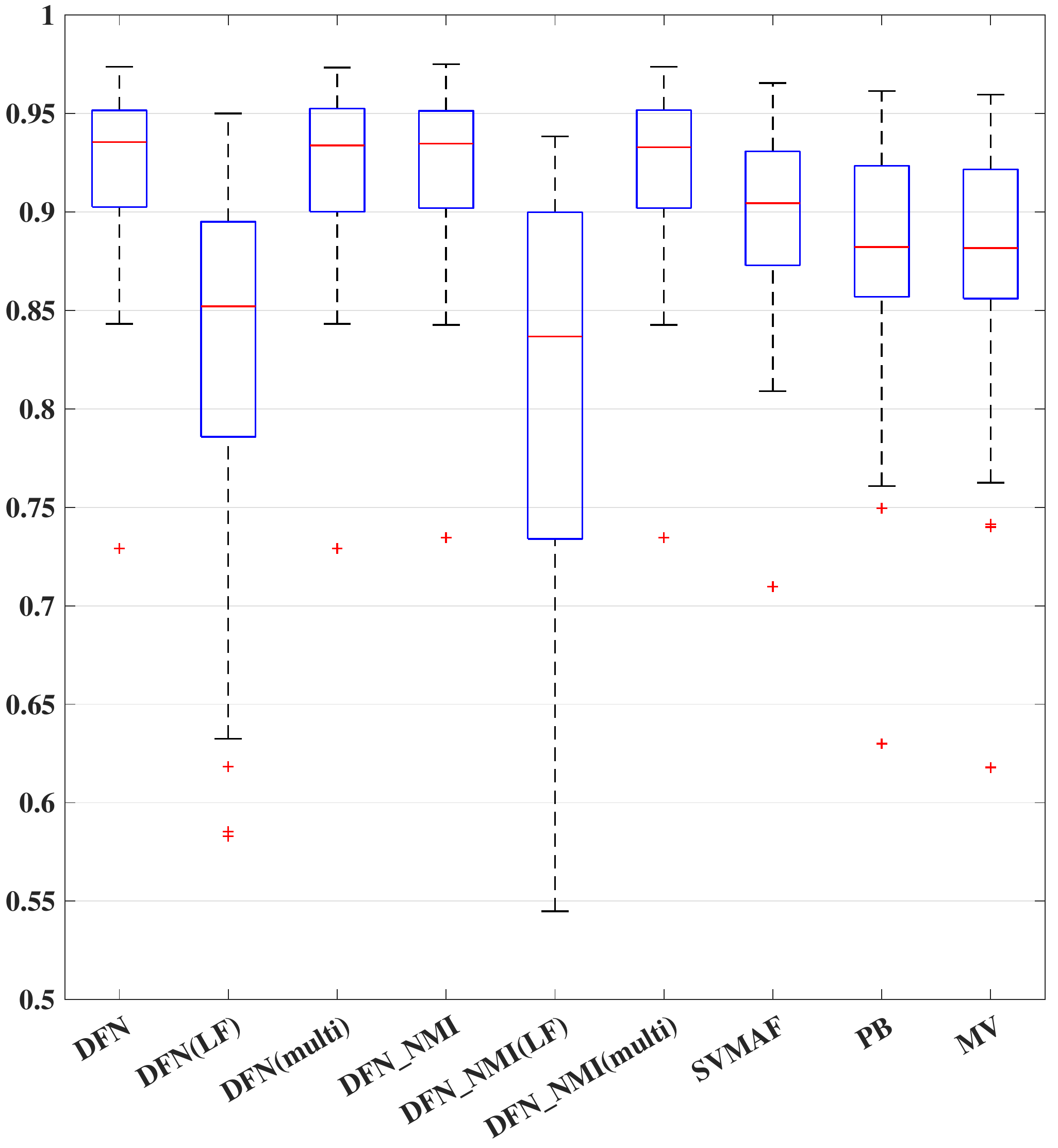}
		\end{minipage}
	}
	\caption{Comparison of (a) {epicardium} and (b) {endocardium} segmentation accuracies for different methods on LV-09 dataset in averaged 3D Dice metric (ADM). The ADM are averaged over all testing subjects. The red central mark of the box is the median and the edges are the 25th and 75th percentiles, while the red plus signs represent outliers.}
	\label{fig:LV09_box}
\end{figure*}

\begin{figure*}[!hbtp]
	\centering
	\setlength{\abovecaptionskip}{2mm}
	\setlength{\belowcaptionskip}{0mm}
	\subfloat[Epicardium segmentation results at ED frame.]{
		\label{fig:LV09o_visual}
		\begin{minipage}[t]{\textwidth}
			\centering
			\includegraphics[width=12.0cm]{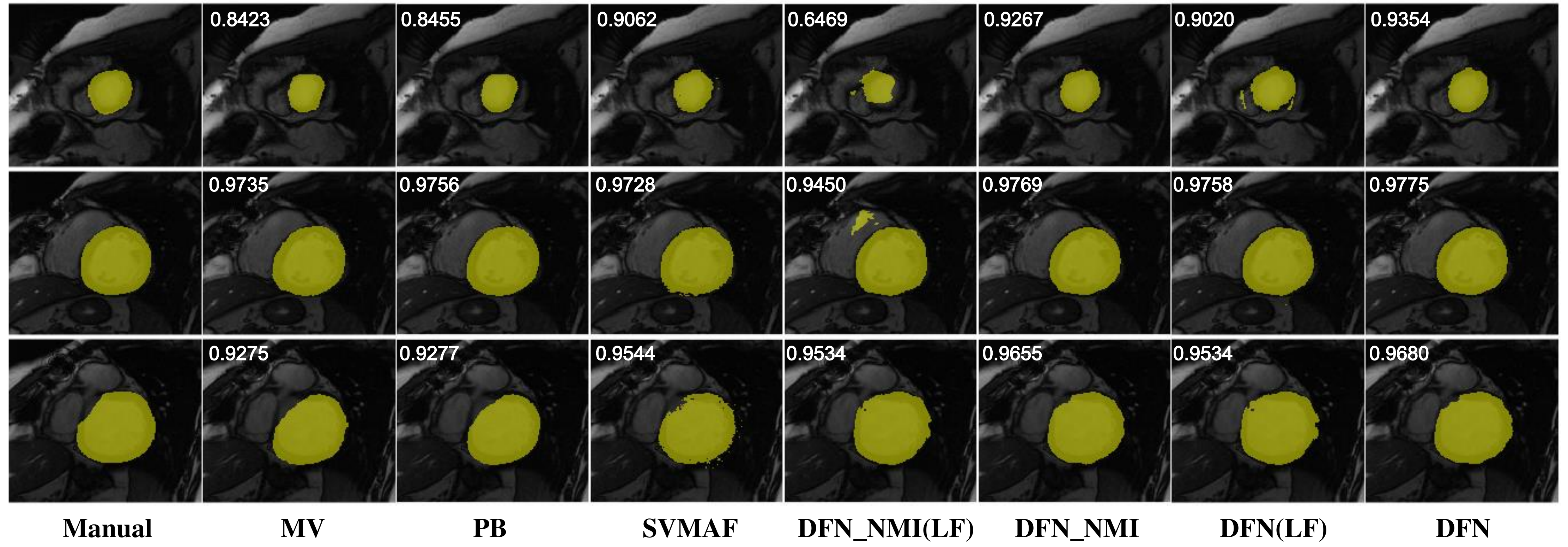}
		\end{minipage}
	} \\
	\subfloat[Endocardium segmentation results at ED frame.]{
		\label{fig:LV09i_visual}
		\begin{minipage}[t]{\textwidth}
			\centering
			\includegraphics[width=12.0cm]{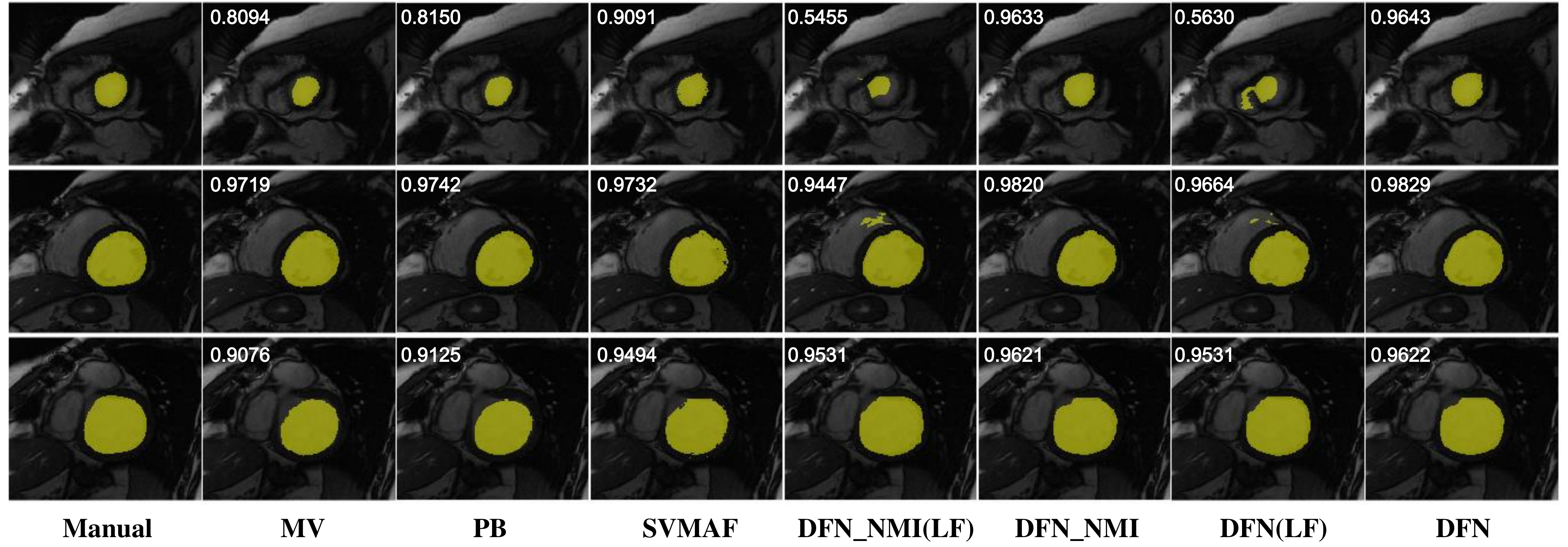}
		\end{minipage}
	}
	\caption{Visual comparison of segmentation results using different methods on LV-09 dataset. Epicardium slices in (a) and  endocardium slices in (b) are from one subject at testing set. Please refer to the electronic version for better comparison. Top to bottom: the basal, mid-ventricular and apical slices from the same subject. Left to right: manual labeling, majority voting (MV), patch-based method (PB), SVM segmentation with augmented features (SVMAF), DFN\_NMI using LF registration, DFN\_NMI, DFN using LF registration, and DFN. The small white text in each sub-image is the corresponding 2D Dice metric {value}.}
	\label{fig:LV09_visual}
\end{figure*}

Our DFN and DFN\_NMI based on LF registration achieve lower values of ``Good" percentage in all comparisons, due to the inaccurate registration results without using landmarks on this dataset. This indicates that our method, as a multi-atlas segmentation method, relies on a relatively good registration method, and breaks down on LV-09 dataset when LF registration does not work well.
Figure~\ref{fig:LV09_box} compares the segmentation accuracies in the box-plot, and Figure~\ref{fig:LV09_visual} shows the epicardium and endocardium segmentation results using different methods for the basal, mid-ventricular and apical slices from one subject in testing set.

\subsection{Evaluation on network architecture}

\subsubsection{Impact of sigmoid layer}
\label{sec:nece_sigmoid_layer}
In the feature extraction subnet discussed in section 2.2, we added a sigmoid layer at the end of {this} subnet {to suppress the magnitudes of deep features for robustness}, and its outputs are taken as voxel-wise features for computing label fusion weights. We now experimentally evaluate the necessity of sigmoid layer on segmentation performance.

First, we train two deep fusion nets with or without sigmoid layer for epicardium at ED frame using 15 training subjects, and test on 30 testing and online subjects of LV-09 database. 
The testing accuracies are reported in Table~\ref{tab:SigmoidLayer}. The results show that the network with sigmoid layer is better than the one without sigmoid layer in ``Good" percentage ($p\mbox{-value} < 0.05$) and APD ($p\mbox{-value} < 0.001$) with comparable ADM, indicating the effectiveness of the sigmoid layer on improving the segmentation accuracies.

\begin{table*}[!bp] \small
	\centering
	\renewcommand\arraystretch{1.4}
	\begin{tabu}{c*{3}{@{\hspace{1.0em}} c}}
		\tabucline[1.5pt]{-}
		Normalization  &``Good" percentage & ADM &  APD \\
		\tabucline[1.2pt]{-}
		DFN (w/ sigmoid) & \textbf{97.59}(\textbf{6.18})   &  \textbf{0.9499}(\textbf{0.01})  &  \textbf{1.97}(\textbf{0.30})     \\
		DFN (w/o sigmoid) & 94.49(11.80)  & 0.9464(0.01)  &  2.15(0.43)  \\
		\tabucline[1.5pt]{-}
	\end{tabu}
	\caption{{Epicardium} segmentation accuracies of deep fusion nets with or without sigmoid layer at ED frame on {testing and online sets} of LV-09 dataset. The number of selected atlases is fixed as 5 for both training and testing. Each cell is formatted as ``mean (standard deviation)".}
	\label{tab:SigmoidLayer}
\end{table*}

Second, we also evaluate the performance of learned deep features with and without sigmoid layer as similarity measure. We compare the majority voting methods respectively using deep features learned by deep fusion nets with sigmoid layer (referred as Deep features (w/ sigmoid)) or without sigmoid layer (referred as Deep features (w/o sigmoid)), and normalized mutual information (referred as NMI) as similarity measures on epicardium segmentation at ED frame of LV-09 database. Only training set is used as atlases, and the testing accuracies are computed in testing and online sets. 
The results are listed in Table~\ref{tab:MV_SigmoidLayer}, showing that deep features learned with sigmoid layer {work} better than the ones learned without sigmoid layer and NMI in ``Good" percentage ($p\mbox{-value} < 0.05$) with comparable ADM and APD.

\begin{table*}[!tp] \small
	\centering
	\renewcommand\arraystretch{1.4}
	\begin{tabu}{c*{3}{@{\hspace{1.0em}} c}}
		\tabucline[1.5pt]{-}
		Similarity   &``Good" percentage & ADM &  APD \\
		\tabucline[1.2pt]{-}
		Deep features (w/ sigmoid) & \textbf{84.39}(\textbf{13.18}) & {0.9056}({0.02})  & \textbf{3.03}(\textbf{0.50}) \\
		Deep features (w/o sigmoid) & 79.37(16.19) & 0.9012(0.02) & 3.25(0.47)  \\
		NMI & 74.65(15.31)  & \textbf{0.9087}(\textbf{0.02}) & 3.07(0.54)  \\
		\tabucline[1.5pt]{-}
	\end{tabu}
	\caption{{Epicardium segmentation accuracies of majority voting using different atlas selection strategies at ED frame on testing and online sets of LV-09 dataset. Atlas selection is implemented respectively using NMI, deep features learned with and without sigmoid layer}. Each cell is formatted as ``mean (standard deviation)".}
	\label{tab:MV_SigmoidLayer}
\end{table*}

\begin{table*}[!bp] \small
	\centering
	\renewcommand\arraystretch{1.4}
	\begin{tabu}{c*{3}{@{\hspace{2em}} c}}
		\tabucline[1.5pt]{-}
		Loss  &``Good" percentage & ADM &  APD \\
		\tabucline[1.2pt]{-}
		$L_2$    & \textbf{97.59}(\textbf{6.18})  & 0.9499(0.01) & \textbf{1.97}(\textbf{0.30}) \\
		$L_1$    & 97.39(9.41)  & \textbf{0.9503}(\textbf{0.01}) & 2.00(0.34) \\
		Hinge & 96.90(8.19)  & 0.9490(0.01) & 2.04(0.36) \\
		Dice  & 96.75(9.27)  & 0.9475(0.01) & 2.03(0.37) \\
		Log   & 95.61(10.25) & 0.9486(0.01) & 2.05(0.34) \\
		\tabucline[1.5pt]{-}
	\end{tabu}
	\caption{{Epicardium} segmentation accuracies of deep fusion nets with different loss layers at ED frame on {testing and online sets} of LV-09 dataset. The number of selected atlases is fixed as 5 for both training and testing. Each cell is formatted as ``mean (standard deviation)".}
	\label{tab:lossLayer}
\end{table*}

\subsubsection{Impact of loss layer}
\label{sec:comp_loss_layer}
In this experiment, we compare {the} performance of our methods using 5 different loss functions, \emph{i.e.}, $L_2$, $L_1$, hinge~\citep{Cortes1995}, Dice~\citep{Milletari2016} and log~\citep{Murphy2012} losses. We train 5 different networks for epicardium at ED using 15 training subjects, and test on 30 testing and online subjects of LV-09 database. The learning rates are tuned such that the loss can decently decrease.
The testing accuracies reported in Table~\ref{tab:lossLayer} show that $L_2$ loss achieves marginally better accuracies compared with some traditional segmentation losses, \emph{e.g.}, hinge, Dice and log losses. We failed to train a converged parameter set for deep fusion net using cross-entropy loss,  which seems to be counter-intuitive. But due to the specially designed linear voting layer before the loss layer, the cross-entropy loss causes unstable gradients for error back-propagation in network training (please refer to section 2.3).

\subsubsection{Impact of search volume}
\label{sec:impact_net_parameters}
We now evaluate the influence of search volume $R_{nl}$, \emph{i.e.}, the non-local region for patch-based label fusion, on the segmentation performance. 
As shown in Table~\ref{tab:impact_SearchVolume}, we respectively train five DFNs with different search volumes, whose sizes range from 1 to 9 with interval of 2, on epicardium at ED frame using 15 training subjects, and test on 30 testing and online subjects of LV-09 dataset. 
The results indicate that the segmentation accuracies are not sensitive to  search volume, and generally a larger size of search volume could produce marginally better segmentation accuracies. This is reasonable since larger search volume provides more patch candidates around registered pixels for label fusion, resulting in robustness to inaccurate registrations.

\begin{table*}[!tp]\small
	\centering
	\renewcommand\arraystretch{1.4}
	\begin{tabu}{c*{3}{@{\hspace{1.5em}} c}}
		\tabucline[1.5pt]{-}
		$R_{nl}$ &``Good" percentage & ADM &  APD        \\
		\tabucline[1.2pt]{-}
		$[1,1]$ &  97.35(6.50)   &  0.9418(0.01)  &  2.17(0.36)     \\
		$[3,3]$ &  96.15(11.72)  &  0.9453(0.01)  &  2.10(0.35)     \\
		$[5,5]$ &  96.03(9.47)   &  0.9483(0.01)  &  1.96(0.30)     \\
		$[7,7]$ &  97.59(6.18)   &  0.9499(0.01)  &  1.97(0.30)     \\
		$[9,9]$ & \textbf{97.89}(\textbf{6.06}) &  \textbf{0.9507}(\textbf{0.01}) & \textbf{1.94}(\textbf{0.32}) \\
		\tabucline[1.5pt]{-}
	\end{tabu}
	\caption{{Epicardium} segmentation accuracies by training DFNs with different sizes of search volumes $R_{nl}$ on training set and testing on {testing and online sets} of LV-09 dataset. The number of selected atlases is fixed as 5 for both training and testing. Each cell is formatted as ``mean (standard deviation)".}
	\label{tab:impact_SearchVolume}
\end{table*}

\subsection{Comparison on atlas selection strategy}
\label{sec:impact_atlas_selec}
In the training and testing phases of our DFNs, we could choose different atlas selection strategies, \emph{e.g.}, selecting different numbers of atlases, using either deep feature distance or NMI for atlas selection. In this section, we will test the impacts of different atlas selection strategies on the performance of DFN.

\begin{figure*}[!htbp]
	\centering
	\setlength{\abovecaptionskip}{2mm}
	\setlength{\belowcaptionskip}{0mm}
	\subfloat[SATA-13 (LB Registration)]{
		\label{fig:AtlasSelec_a}
		\begin{minipage}[t]{0.48\textwidth}
			\centering
			\includegraphics[width=6.0cm]{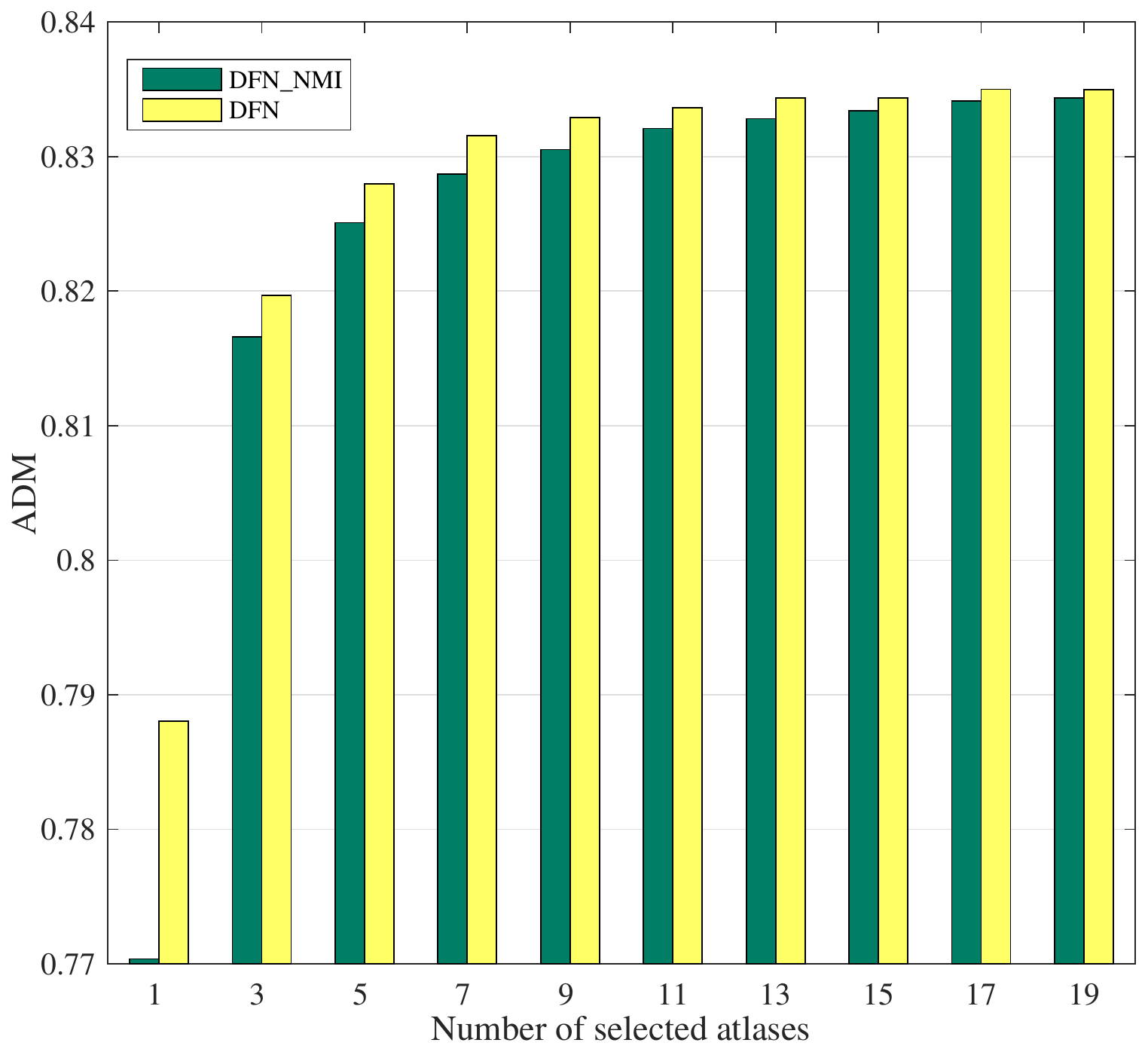}
		\end{minipage}
	}
	\subfloat[SATA-13 (LF Registration)]{
		\label{fig:AtlasSelec_b}
		\begin{minipage}[t]{0.48\textwidth}
			\centering
			\includegraphics[width=6.0cm]{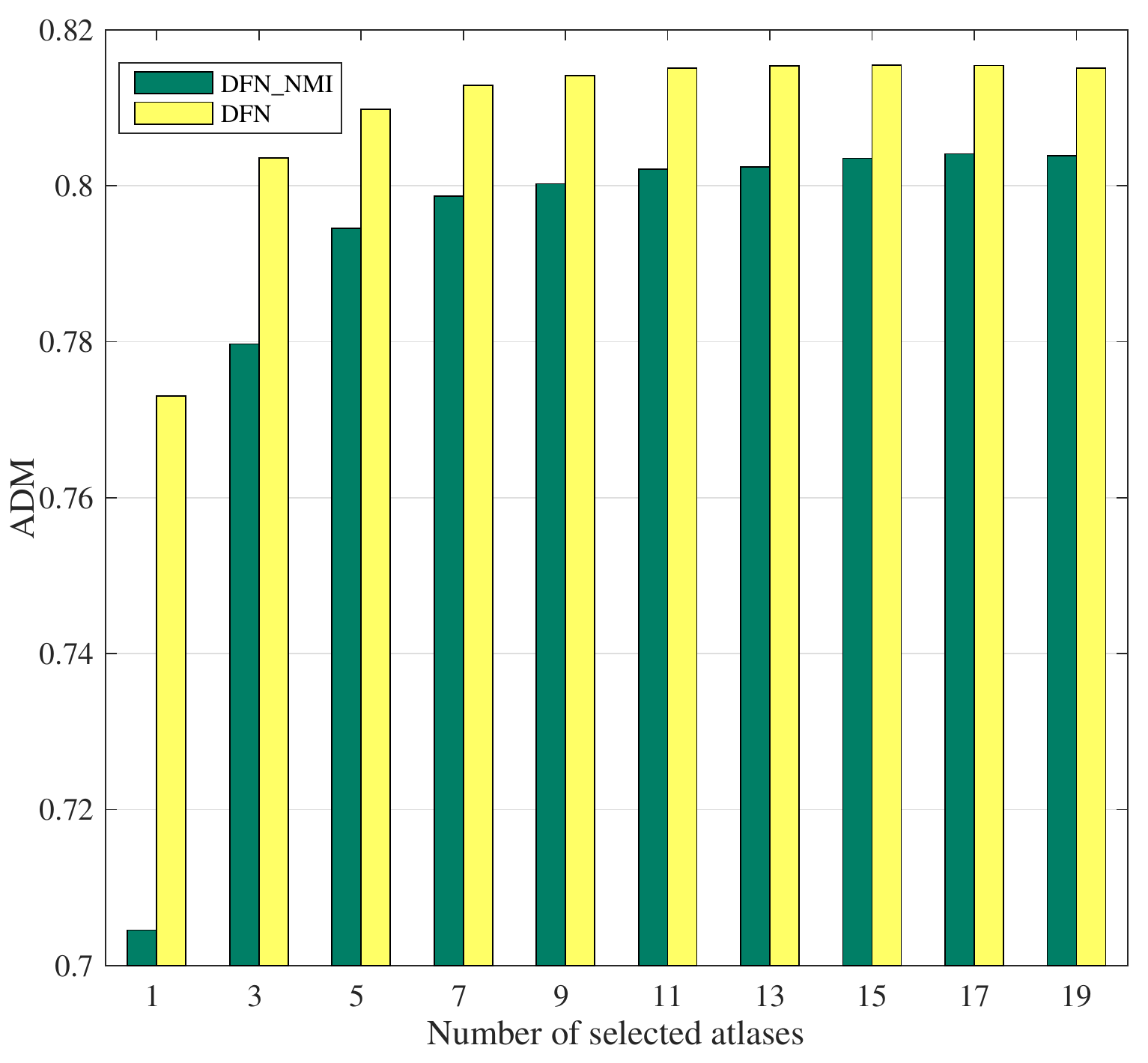}
		\end{minipage}
	}
	
	\subfloat[LV-09 (Epicardium)]{
		\label{fig:AtlasSelec_c}
		\begin{minipage}[t]{0.48\textwidth}
			\centering
			\includegraphics[width=6.0cm]{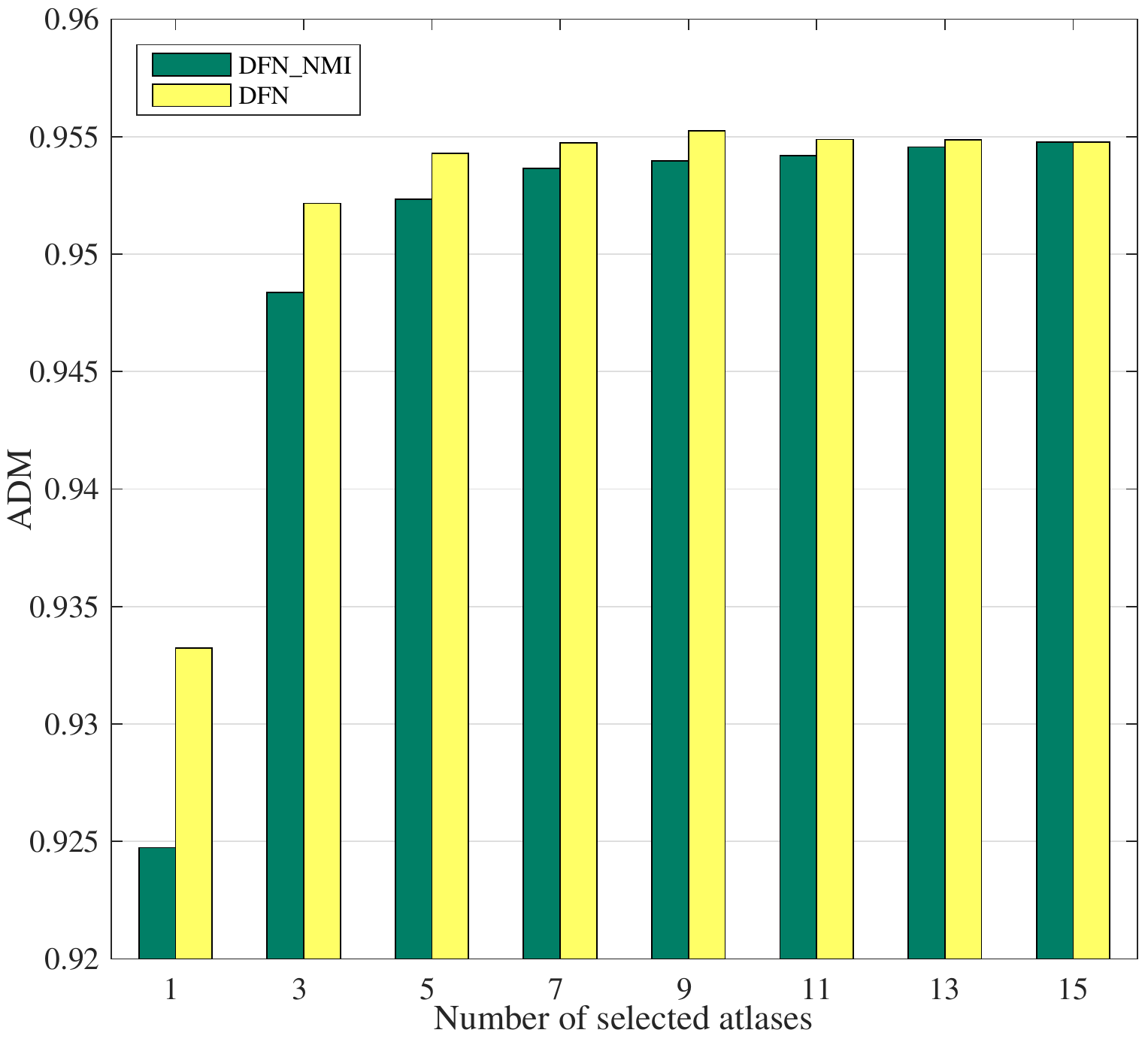}
		\end{minipage}
	}
	\subfloat[LV-09 (Endocardium)]{
		\label{fig:AtlasSelec_d}
		\begin{minipage}[t]{0.48\textwidth}
			\centering
			\includegraphics[width=6.0cm]{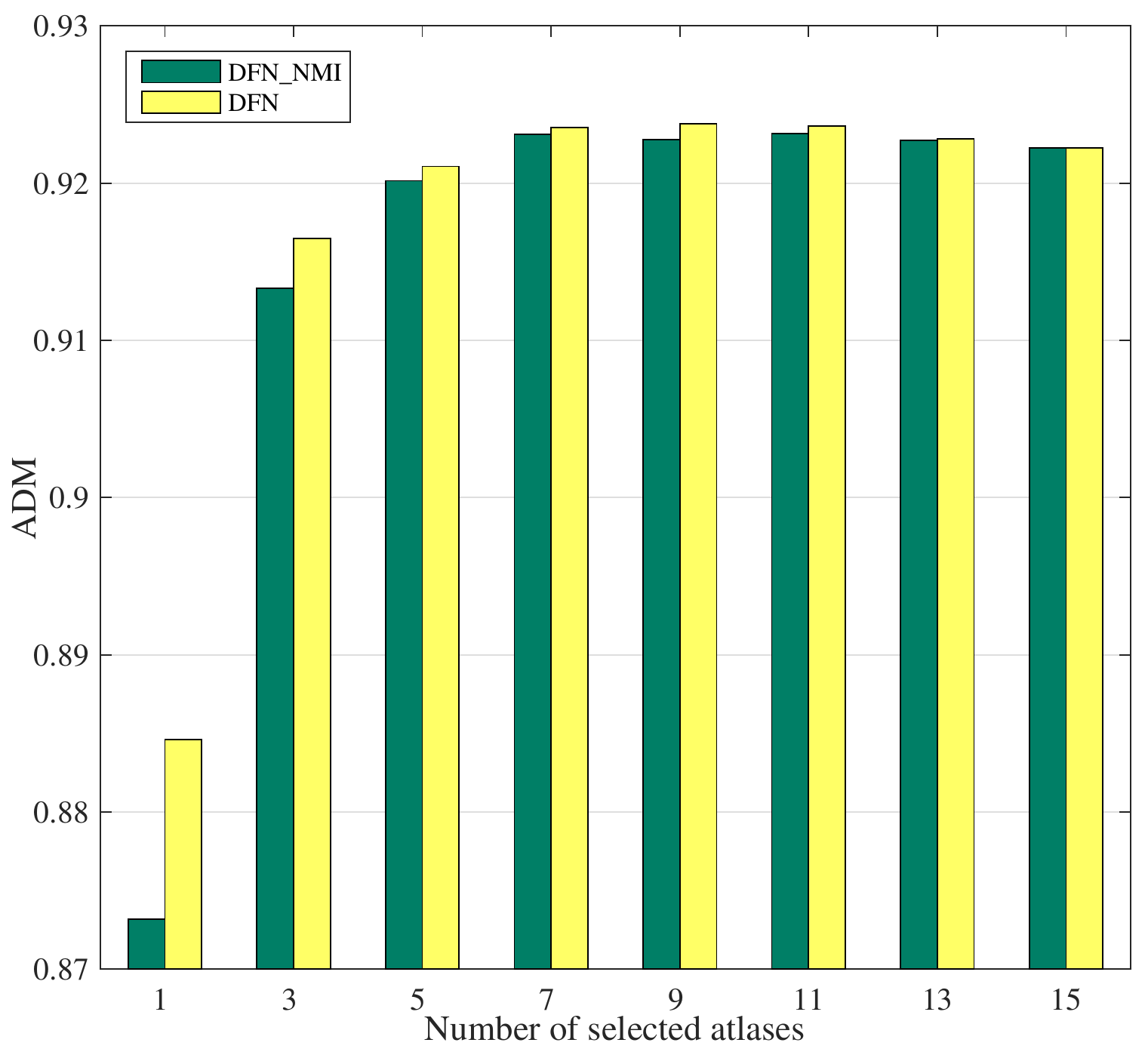}
		\end{minipage}
	}
	\caption{The change of averaged 3D Dice metric (ADM) \emph{w.r.t.} the number of selected atlases at testing process. {The ADM are averaged over all testing subjects.} (a) and (b) are respectively {using LB registration and LF registration for myocardium segmentation} on SATA-13 dataset, while (c) and (d) are for epicardium and endocardium segmentations utilizing LB registration on LV-09 dataset.}
	\label{fig:atlasSelec}
\end{figure*}

We first compare the segmentation accuracies of our proposed methods (referred to DFN\_NMI and DFN) with respect to different numbers of selected atlases at testing phase. The deep fusion nets learned in sections~\ref{sec:SATA13_accuracy} and \ref{sec:LV09_accuracy}, which use randomly selected 5 atlases according to a distribution proportional to NMI between target image and atlases at training phase, are utilized in this comparison. 
As shown in Fig.~\ref{fig:atlasSelec}, we compare the performance of DFN\_NMI and DFN on SATA-13 dataset (Fig.~\ref{fig:AtlasSelec_a} and \ref{fig:AtlasSelec_b}) and LV-09 dataset (Fig.~\ref{fig:AtlasSelec_c} and \ref{fig:AtlasSelec_d}) respectively. 
The experimental results show that atlas selection using deep feature distance consistently works better than that using NMI for different numbers of selected atlases at testing phase. Moreover, using larger {numbers} of atlases generally {produces} better ADM scores, but the accuracies saturate after around 11 atlases.

\begin{table*}[!tp]\small
	\centering
	\renewcommand\arraystretch{1.4}
	\begin{tabu}{c*{4}{@{\hspace{1em}} c}}
		\tabucline[1.5pt]{-}
		Training & Testing & ``Good" percentage & ADM & APD \\
		\tabucline[1.2pt]{-}
		\multirow{2}*{Random5\_NMI} &  Top5\_DF  & \textbf{97.59}(\textbf{6.18})  &  0.9499(0.01)  &  \textbf{1.97}(\textbf{0.30})  \\
		~                           & Top5\_NMI  & 95.71(9.34)  &  \textbf{0.9503}(\textbf{0.01})  &  2.01(0.30)  \\
		\midrule
		\multirow{2}*{Top5\_DF}     &  Top5\_DF  & 97.40(6.46)  & 0.9483(0.01)  &  2.00(0.33)  \\
		~                           & Top5\_NMI  & 94.59(10.26) & 0.9488(0.01)  &  2.03(0.29)  \\
		\midrule
		\multirow{2}*{Top5\_NMI}    &  Top5\_DF  & 96.90(8.99)  & 0.9468(0.01)  &  2.02(0.32)  \\
		~                           & Top5\_NMI  & 95.30(10.51) & 0.9472(0.01)  &  2.05(0.33)  \\
		\tabucline[1.5pt]{-}
	\end{tabu}
	\caption{{Epicardium} segmentation accuracies of DFNs with different atlas selection strategies at training and testing phases on {testing and online sets} of LV-09 dataset, where only training set is used as training data. Each cell is formatted as ``mean (standard deviation)".}
	\label{tab:impact_AtlasSelecStrategy}
\end{table*}

We next experimentally investigate the advantages of our random atlas selection strategy at training phase, which has been discussed in section~\ref{network_training}.  By fixing the number of selected atlases to 5, we compare the performance of DFNs using three different atlas selection strategies in each forward-backward computation of training process, \emph{i.e.}, selecting top-5 atlases with smallest deep feature distances (top5\_DF), selecting top-5 atlases with largest normalize mutual information (top5\_NMI), and randomly selecting atlases according to a distribution proportional to NMI between target and atlas images (random5\_NMI).  We train DFNs with the above strategies for epicardium at ED using 15 training subjects, and test on 30 testing and online subjects of LV-09 dataset.
\begin{table*}[!bp]\small
	\centering
	\renewcommand\arraystretch{1.4}
	\begin{tabu}{c*{3}{@{\hspace{1.5em}} c}}
		\tabucline[1.5pt]{-}
		$K_0$ &``Good" percentage & ADM &  APD \\
		\tabucline[1.2pt]{-}
		$1$ &  95.66(9.60)   &   0.9335(0.02)   &   2.34(0.50)    \\
		$3$ &  97.17(6.43)   &   0.9463(0.01)   &   2.07(0.37)    \\
		$5$ &  \textbf{97.59}(\textbf{6.18})   &   \textbf{0.9499}(\textbf{0.01})   &   1.97(0.30)    \\
		$7$ &  97.14(8.78)   &   \textbf{0.9499}(\textbf{0.01})   &   \textbf{1.93}(\textbf{0.29})    \\
		$9$ &  96.25(10.00)  &   0.9496(0.01)   &   1.97(0.29)    \\
		\tabucline[1.5pt]{-}
	\end{tabu}
	\caption{{Epicardium} segmentation accuracies by training DFNs with different numbers of selected atlases $K_0$ on training set and testing on {testing and online sets} of LV-09 dataset. The size of search volume is fixed as [7,7], and the number of selected atlases for testing is same as the corresponding training one. Each cell is formatted as ``mean (standard deviation)".}
	\label{tab:impact_Training Atlas}
\end{table*}
For each strategy at training phase, we attempt two different atlas selection strategies at testing phase, \emph{i.e.}, select top-5 atlases with largest normalize mutual information (top5\_NMI) and select top-5 atlases with smallest deep feature distance (top5\_DF). The experimental results are listed in Table~\ref{tab:impact_AtlasSelecStrategy}. Compared with the strategies of top5\_DF and top5\_NMI at training phase, the strategy of using random5\_NMI performs best in all metrics.  Moreover, random5\_NMI paired with top5\_DF strategies at training and testing phases achieves the best accuracies among all the compared strategies.

In Table~\ref{tab:impact_Training Atlas}, we also compare the performance of DFN by varying the numbers of selected atlases (denoted as $K_0$) at training phase using random atlas selection strategy, with $K_0$ atlases selected by deep feature distance at testing phase. The experimental results show that the segmentation accuracies are relatively stable to $K_0$, \emph{e.g.}, the accuracies are not significantly lower even when using 1 training atlas (\emph{i.e.}, $K_0 = 1$). This is probably because the random atlas selection strategy enforces that each target image at training phase can be paired with diverse atlases. Empirically, $K_0$ with values of 5 to 7 produces best accuracies.

\subsection{Cross-dataset evaluation}
To evaluate the generalization abilities of the learned DFNs across different datasets, we conduct experimental comparisons for network training and testing across two different datasets. More specifically, we train each DFN on one dataset and test it on the other one. 
When we test the DFN learned from training dataset on the testing dataset, we attempt two different atlas selection strategies for the testing subjects, \emph{i.e.}, selecting atlases from the subjects in the training dataset or in the testing dataset.
We create the segmentation mask of myocardium for LV-09 dataset by subtracting the endocardium mask from the epicardium mask, to be compatible with the ground-truth segmentation mask provided in SATA-13 dataset.
The experiments are performed on ED frame of both datasets, and the accuracy is evaluated by averaged 3D Dice metric (ADM) and averaged 3D Hausdorff distance (AHD) over all subjects at ED frame of the testing dataset.

\begin{table*}[!bp]\small
	\centering
	\renewcommand\arraystretch{1.4}
	\begin{tabu}{l*{2}{@{\hspace{1em}} c}}
		\tabucline[1.5pt]{-}
		Method       & Averaged Dice metric & Averaged Hausdorff distance\\
		\tabucline[1.2pt]{-}
		MV           &  0.620(0.066)        &   16.58(3.22)              \\
		PB           &  0.639(0.067)        &   16.69(3.26)              \\
		SVMAF        &  0.691(0.082)        &   22.15(5.82)              \\
		\hdashline
		DFN          & \textbf{0.734}(\textbf{0.058}) & \textbf{15.32}(\textbf{3.88}) \\  
		DFN\_NMI     &  0.730(0.060)        &   15.82(4.44)              \\
		\tabucline[1.5pt]{-}
	\end{tabu}
	\caption{{Myocardium} segmentation results by {training DFNs on LV-09 dataset} and {testing on SATA-13 dataset}, where 45 subjects of LV-09 dataset are used as atlases in the testing process. Each cell is formatted as ``mean (standard deviation)".}
	\label{tab:SATA13_CrossDB_acc}
\end{table*}

\begin{figure*}[!bp]
	\setlength{\abovecaptionskip}{2mm}
	\setlength{\belowcaptionskip}{0mm}
	\subfloat[Testing on SATA-13 dataset]{
		\label{fig:visual_crossDB_TrainOnLV09}
		\begin{minipage}[t]{0.48\textwidth}
			\centering
			\includegraphics[width=6.0cm]{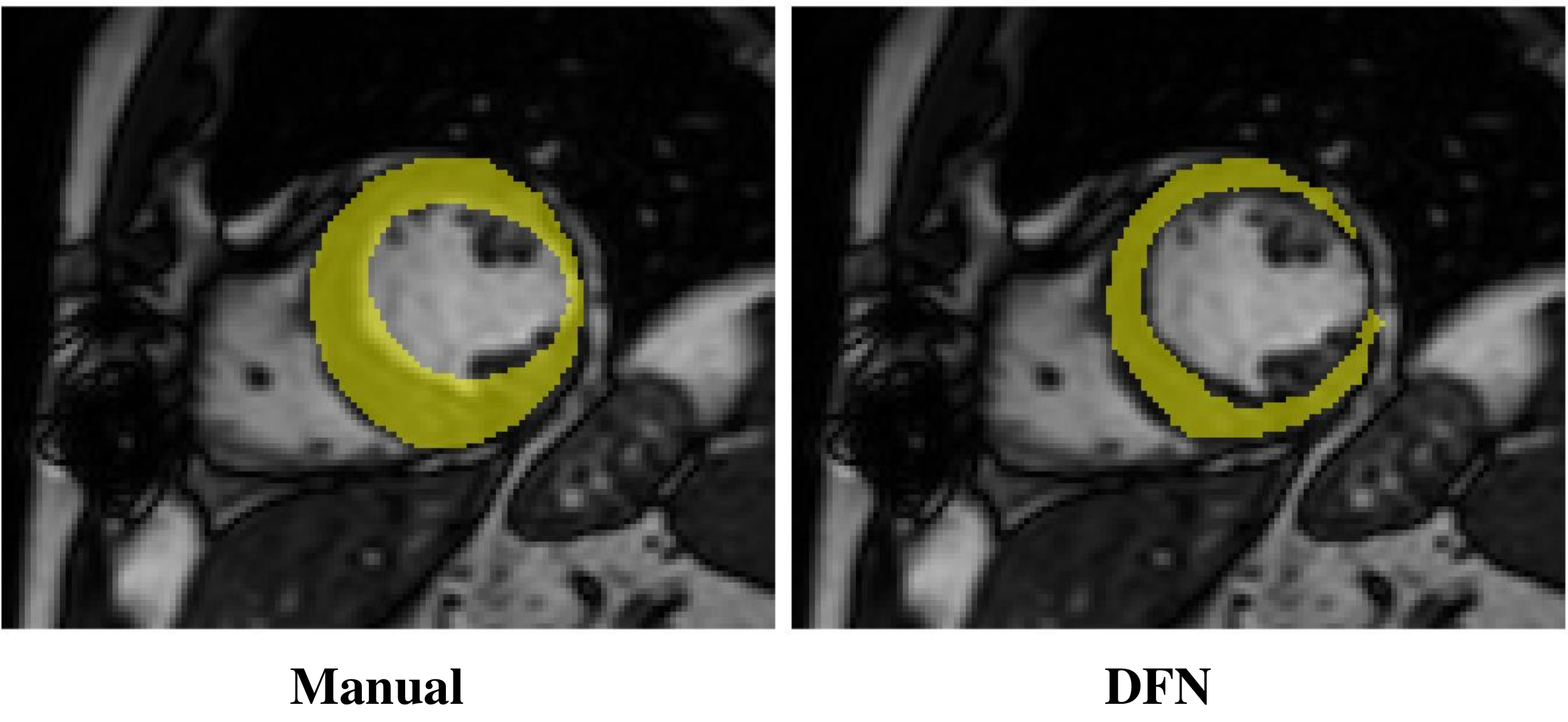}
		\end{minipage}
	}
	\subfloat[Testing on LV-09 dataset]{
		\label{fig:visual_crossDB_TrainOnSATA13}
		\begin{minipage}[t]{0.48\textwidth}
			\centering
			\includegraphics[width=6.0cm]{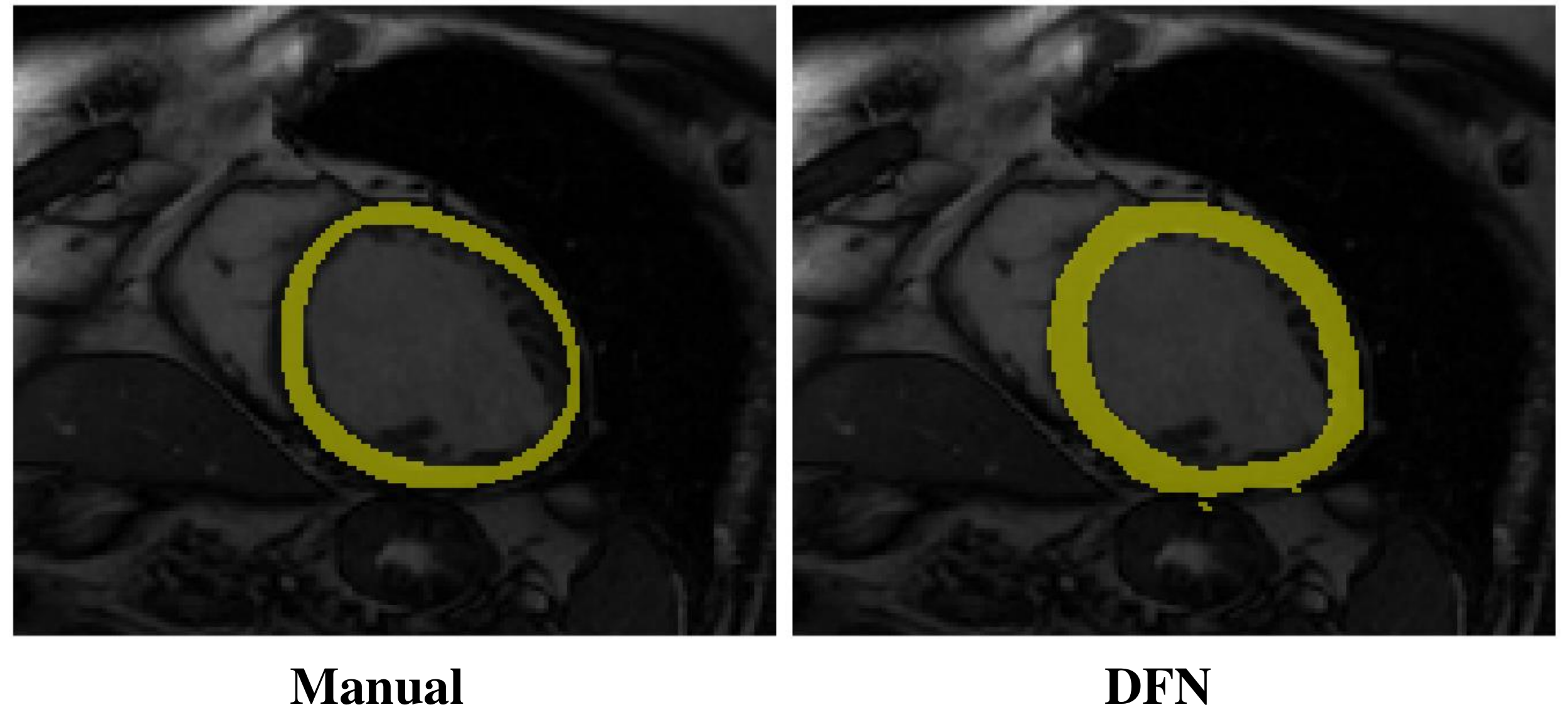}
		\end{minipage}
	}
	\caption{Visual examples of estimated myocardium segmentations by DFN (right)  and corresponding manual segmentations (left) in (a) SATA-13 dataset and (b) LV-09 dataset. As we can see, manual segmentation in (a) SATA-13 dataset is much ``thicker" than the one in (b) LV-09 dataset. In (b) LV-09 dataset, the estimated segmentation by DFN is much ``thicker" than the manual one, possibly because the network here is trained on ``thick" labels of SATA-13 dataset and utilizes ``thick" atlases to estimate the target label. Similarly, the estimated segmentation in (a) SATA-13 dataset is much ``thinner" than the manual one.}
	\label{fig:visual_crossDB}
\end{figure*}

\begin{figure*}[!tp]
	\setlength{\abovecaptionskip}{2mm}
	\setlength{\belowcaptionskip}{0mm}
	\subfloat[Testing accuracy on SATA-13 dataset]{
		\label{fig:crossDB_TrainOnLV09}
		\begin{minipage}[t]{0.48\textwidth}
			\centering
			\includegraphics[width=6.0cm]{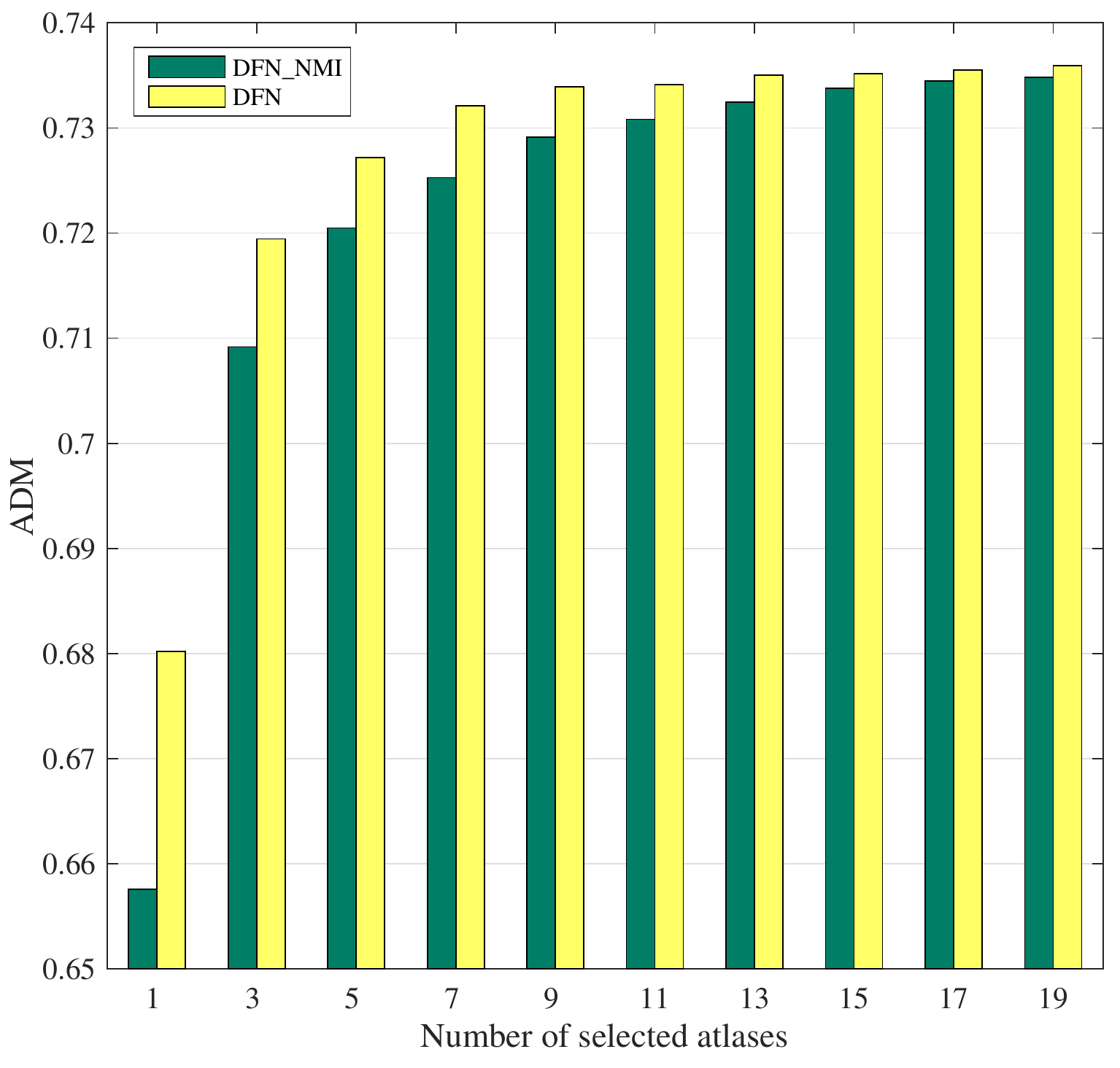}
		\end{minipage}
	}
	\subfloat[Testing accuracy on LV-09 dataset]{
		\label{fig:crossDB_TrainOnSATA13}
		\begin{minipage}[t]{0.48\textwidth}
			\centering
			\includegraphics[width=6.0cm]{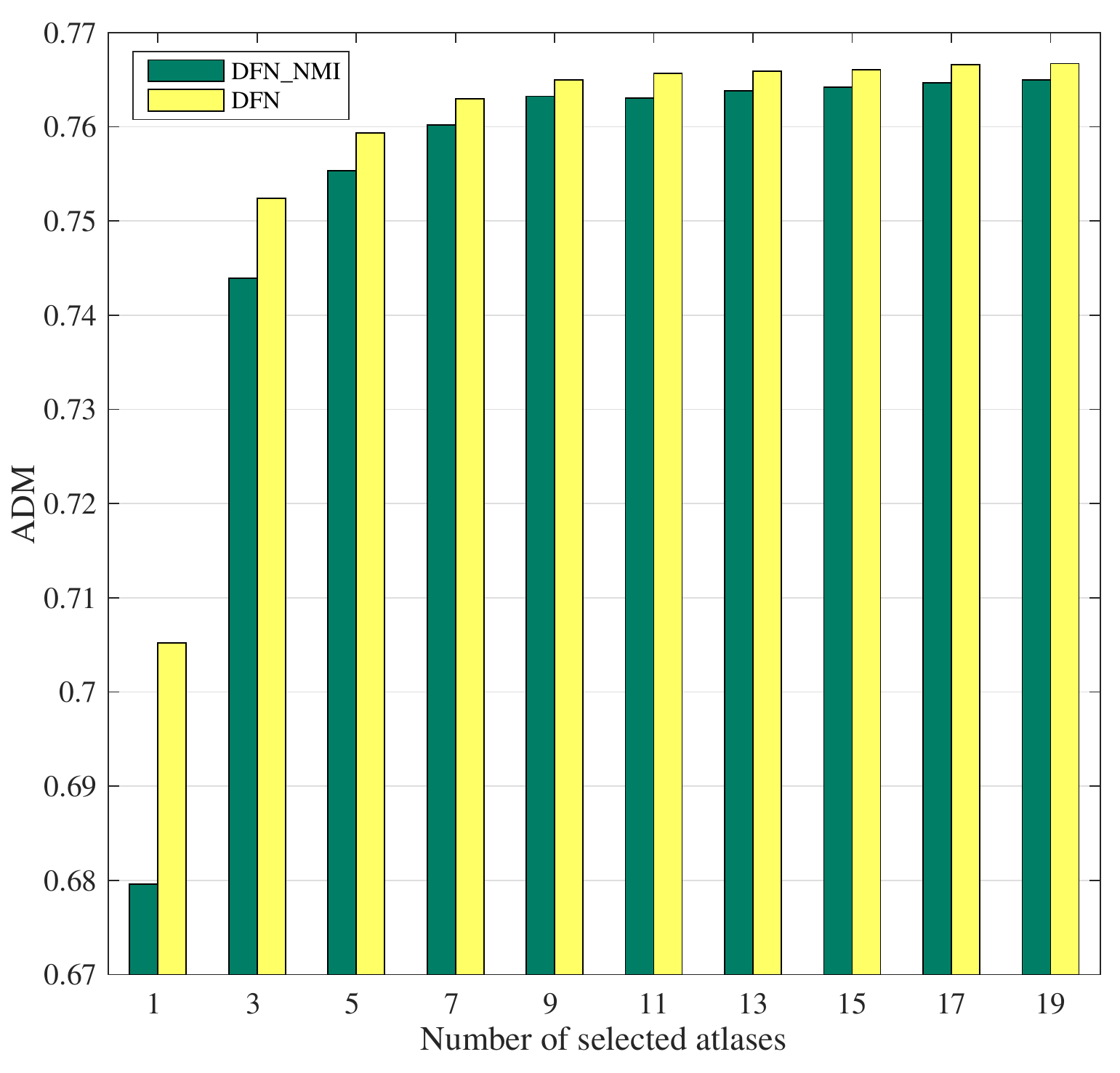}
		\end{minipage}
	}
	\caption{The change of averaged 3D Dice metric (ADM) \emph{w.r.t.} the number of selected atlases respectively on (a) SATA-13 dataset and (b) LV-09 dataset for myocardium segmentation at testing phase. {The ADM are averaged over all testing subjects.}}
	\label{fig:crossDB_AtlasSelec}
\end{figure*}

We first evaluate the cross-dataset performance using the strategy that selects atlases from the subjects in the training dataset for the testing subjects in the testing dataset,  and the segmentation accuracies are listed in Table~\ref{tab:SATA13_CrossDB_acc} (train DFN on LV-09 and test it on SATA-13) and Table~\ref{tab:LV09_CrossDB_acc}  (train DFN on SATA-13 and test it on LV-09). The experimental results show that our learned DFNs achieve significantly higher ADMs ($p\mbox{-value} < 0.001$) and marginally better AHDs than traditional multi-atlas segmentation methods on both two datasets.
Compared with the methods that perform training and testing on the same dataset, \emph{e.g.}, Table~\ref{tab:SATA13_acc} in section~\ref{sec:SATA13_accuracy}, the accuracies of our DFN for cross-dataset evaluation are relatively low, which is possibly because of the different styles of manual labels in these two datasets, as shown in Fig.~\ref{fig:visual_crossDB} (please refer to the figure for detailed descriptions of this difference).
Figure~\ref{fig:crossDB_AtlasSelec} shows the performance of deep fusion nets using different numbers of selected atlases respectively on two datasets in the testing process, indicating that our defined deep feature distance consistently works better than NMI in all metrics.

\begin{table*}[!tp]\small
	\centering
	\renewcommand\arraystretch{1.4}
	\begin{tabu}{l*{2}{@{\hspace{1em}} c}}
		\tabucline[1.5pt]{-}
		Method       & Averaged Dice metric & Averaged Hausdorff distance\\
		\tabucline[1.5pt]{-}
		MV           &  0.672(0.057)        &    11.83(3.10)             \\
		PB           &  0.681(0.055)        &    11.45(2.55)             \\
		SVMAF        &  0.708(0.044)        &    17.63(4.93)             \\
		\hdashline
		DFN          & \textbf{0.766}(\textbf{0.042}) & \textbf{10.68}(\textbf{3.84}) \\  
		DFN\_NMI     &  0.763(0.044)        &    11.95(4.78)             \\
		\tabucline[1.5pt]{-}
	\end{tabu}
	\caption{{Myocardium} segmentation results by {training DFNs on SATA-13 dataset} and {testing on LV-09 dataset}, where 83 subjects of SATA-13 dataset are used as atlases in the testing process. Each cell is formatted as ``mean (standard deviation)".}
	\label{tab:LV09_CrossDB_acc}
\end{table*}

We also evaluate the cross-dataset performance using the strategy that selects atlases from the subjects in the testing dataset for testing subjects. In Table~\ref{tab:SATA13_acc}, we present the results of our methods (denoted as DFN(crossDS) and DFN\_NMI(crossDS)) by applying the DFN learned from LV-09 dataset to the SATA-13 dataset without fine-tuning. These accuracies are calculated using the same 5-fold cross validation utilized in section~\ref{sec:SATA13_accuracy}. 
The experimental results show that, if the subjects in SATA-13 dataset are used as atlases for testing subjects, the DFN learned from LV-09 dataset can achieve comparable results on SATA-13 dataset in all metrics, compared with those segmentation methods training and testing on the same SATA-13 dataset. This justifies that our learned feature extraction subnet has well generalization ability across different datasets for LV segmentation.

\section{Conclusions}
In this work, we accomplish the multi-atlas based LV segmentation by a specially designed convolutional neural network. Our network relies on atlas-to-target image registration, and aims to extract  deep features for optimally fusing the warped atlas labels in a non-local patch-based label fusion framework.  This deep fusion net naturally bridges the traditional registration-based multi-atlas approach and modern deep learning approach, and provides a novel deep architecture for solving the tasks of label fusion and atlas selection in multi-atlas segmentation approach.
The proposed net was evaluated SATA-13 and LV-09 datasets for LV segmentation, and  the results demonstrate that it achieves better accuracies in various metrics than the other LV segmentation methods on both datasets, and the only method surpassing ours is the deep learning method using a strong manual prior~\citep{Ngo2017}. We also extensively evaluate the performance of deep fusion nets using variants of architecture, training loss, atlas selection strategy, cross-dataset training and testing, \emph{etc}.

As a registration-based multi-atlas segmentation method, our deep fusion net relies on a good image registration method, and may fail when the atlas-to-target image registration is not accurate enough. For example, {our deep fusion net using landmark-free registration works well on SATA-13 dataset while producing unsatisfactory results on LV-09 dataset}. To improve its robustness to registration errors, first, larger search volume can be utilized to incorporate more voxels around registered voxel for label fusion. Second, it is interesting to build a more robust label fusion subnet for multi-atlas segmentation {via the statistical fusion strategy}.
Moreover, our present study is based on 2D slices mainly due to the constraint of GPU memory.  This is also a common issue for deep learning approaches when applied to 3D medical images.
One common solution is to train the networks using 3D patches instead of full 3D images~\citep{Kamnitsas2017,Dou2017}.

In the future work, we are interested in improving the label fusion subnet by investigating more robust statistical {fusion} strategy, and applying the proposed framework to 3D images.  As a general framework, our deep fusion net can also be applied to other multi-atlas based applications, \emph{e.g.}, image synthesis~\citep{Roy2013}, brain segmentation~\citep{Asman2015}, \emph{etc}.

\section*{Acknowledgments}
This work was supported by the National Natural Science Foundation of China (NSFC) under Grant No. 11622106, 61472313, 11690011 and 61721002, International Exchange Foundation of China NSFC and United Kingdom RS under grant No. 61711530242.

\section*{References}
\bibliography{mybibfile}

\end{document}